\RequirePackage{fix-cm}
\documentclass[smallextended]{svjour3}       
\smartqed  
\usepackage{graphicx}

\usepackage{url}
\usepackage{xcolor}
\usepackage{graphicx}
\usepackage{subfigure}
\usepackage{algorithm}
\usepackage{algorithmicx}
\usepackage{algpseudocode}
\usepackage{amsmath}
\usepackage{setspace}
\usepackage{multirow}
\usepackage[marginal]{footmisc}
\usepackage{url}
\usepackage{balance}
\usepackage[square,sort,comma,numbers]{natbib}
\usepackage{soul}
\usepackage{color}
\usepackage[colorlinks=true,urlcolor=blue,citecolor=blue]{hyperref}
\usepackage{geometry}
\begin{document}

\title{Information Bottleneck Theory on Convolutional Neural Networks
}


\author{JunjieLi         \and
        DingLiu 
}


\institute{JunjieLi \at
              Department of Computer Science and Technology, School of Computer Science and Technology, Tiangong University, Tianjin 300387, China 
           \and
           DingLiu \at
              Department of Computer Science and Technology, School of Computer Science and Technology, Tiangong University, Tianjin 300387, China
              \email{liuding@tiangong.edu.cn}
}

\date{Received: date / Accepted: date}

\maketitle

\begin{abstract}
Recent years, many researches attempt to open the black box of deep neural networks and propose a various of theories to understand it. Among them, Information Bottleneck (IB) theory claims that there are two distinct phases consisting of fitting phase and compression phase in the course of training. This statement attracts many attentions since its success in explaining the inner behavior of feedforward neural networks. In this paper, we employ IB theory to understand the dynamic behavior of convolutional neural networks (CNNs) and investigate how the fundamental features such as convolutional layer width, kernel size, network depth, pooling layers and multi-fully connected layer have impact on the performance of CNNs. In particular, through a series of experimental analysis on benchmark of MNIST and Fashion-MNIST, we demonstrate that the compression phase is not observed in all these cases. This shows us the CNNs have a rather complicated behavior than feedforward neural networks.
\end{abstract}

\section{Introduction}
\label{intro}
In recent, the practical successes of deep neural networks have generated many attempts to explain the performance of deep learning ~\citep{guidotti2019survey,kadmon2016optimal, yu2020learning}, especially in terms of the dynamics of the optimization \citep{saxe2014exact, advani2017high}. In this context, the Information Bottleneck (IB) theory provides a fundamental tool on this topic, and some preliminary empirical exploration of these ideas in deep feedforward neural networks has yielded striking findings~\citep{tishby2015deep, shwartz2017opening, amjad2019learning,poole2019variational,goldfeld2018estimating}. Besides, people start to investigate IB theory from different perspective such as gaussian variables~\citep{chechik2005information, painsky2017gaussian}, multivariate systems~\citep{friedman2013multivariate}, hidden variable networks~\citep{elidan2005learning}, nonlinear systems~\citep{kolchinsky2019nonlinear}, neural network-based classification~\citep{amjad2019learning}, compression of neural networks~\citep{dai2018compressing}, even for the application of document clustering~\citep{slonim2000document} and video search reranking~\citep{hsu2006video}. Moreover, Ref.~\citep{strouse2017deterministic, gabrie2018entropy} discuss the entropy and mutual information in more detail, and Ref.~\citep{shamir2010learning} proves several finite sample bounds which show that information bottleneck can provide representations with good generalization. In addition, Ref. \citep{jonsson2019convergence} presents Mutual Information Neural Estimation (MINE) that can stabilize the test accuracy and reduce its variance.

Inspired from these works, we investigate the IB theory using an analytical method on Convolutional Neural Networks (CNNs), which have wide application on such as image processing in recent years~\citep{wang2019weakly, wang2018getnet}. 
And we observe quite different behaviors in contrast to those of feedforward neural networks. In the series of original works ~\citep{tishby2000information, tishby2015deep, shwartz2017opening}, authors hold some core points that the distinct phases of the Stochastic Gradient Descent (SGD) optimization, drift and diffusion, which explain the empirical error minimization and the representation compression trajectories of
the layers. These phases are characterized by very different signal to noise ratios of the stochastic gradients in every layer. This funding opens the black box of deep learning from the perspective of information theory and draw many attentions. Along this way, a further research offers some different views of IB theory and shows us some different behaviors on feedforward neural networks \citep{saxe2019information}. They say that ``fitting'' and ``compression'' phases in the course of training strongly depend on the nonlinear activation. The authors state that double saturating nonlinearities lead to compression and stochasticity in the training phase does not contribute to compression. Obviously, it is partly in contradiction with the initial idea in Ref.~\citep{shwartz2017opening}. Moreover, Ref.~\citep{gabrie2018entropy} claims that the compression can happen even when using ReLu activation in their high dimensional experiments, and there is not a clear link between compression and generalization. Then lately, some works start to focus on exploring the inner organization of CNNs and autoencoders by using matrix-based Renyi's entropy~\citep{yu2020understanding, yu2019understanding}. The authors propose that variability in the compression behavior strongly depends on different estimators. By using matrix-based Renyi's entropy estimator and remove the redundant information in the MI, they observe compression phase during the training. Moreover, Ref.~\citep{goldfeld2019estimating} find a new phenomenon -- clustering emerging in the training phase. And they propose that the compression strongly rely on the clustering and may not causally related to generalization. So until now, based on all these previous works, compression and the relationship between it and generalization still remain elusive.

In this paper, different from the previous works, we observe no compression phase both on convolution layers and fully connected layers on standard CNNs, even with double saturating nonlinearity such as $tanh$. This observation partly supports the conclusion by Ref.~\citep{saxe2019information} that compression is not the universal phase during the course of training. Moreover, from the perspective of IB theory, we investigate how the fundamental features such as convolutional layer width, network depth, kernel size, pooling layers etc. have an effect on the performance of CNNs. The experimental results verify the importance of these features in improving the generalization performance.

\section{Method}
	
	The Information Bottleneck (IB) theory is introduced by Tishby et.al first time in the paper~\citep{tishby2000information}. Afterwards, Ref.~\citep{tishby2015deep,shwartz2017opening} analyse the training phase of Deep Neural Networks (DNNs) from the perspective of IB. Accordingly, IB suggests that each hidden layer will capture more useful information from the input variable, and the hidden layers are supposed to be the maximally compressed mappings of the input. There are several fundamental points to know about IB theory as follow:

	\subsection{Mutual Information}
	
	Mutual Information (MI) measures the mutual dependence of two random variables. Further, it quantifies the amount of information got about one random variable through observing the other. For example, given two variables $A$ and $B$, mutual information $I(A;B)$ is defined as:
	\begin{equation}
	\begin{aligned}
	I(A;B) &= H(A)-H(A|B)
	\label{eq:I_A_B}
	\end{aligned}
	\end{equation}
	
	\begin{equation}
	\begin{aligned}
	H(A)&=-\sum_{a\in A}p(a)\log p(a)
	\label{eq:H_A}
	\end{aligned}
	\end{equation}
	
	\begin{equation}
	\begin{aligned}
	H(A|B)&=-\sum_{b\in B}p(b)\sum_{a\in A}p(a|b)\log p(a|b)
	\\&=-\sum_{a\in A}\sum_{b \in B}p(b,a)\log p(a|b)
	\label{eq:H_A_B}
	\end{aligned}
	\end{equation}
	
	where $H(A)$ and $H(A|B)$ are entropy and conditional entropy respectively, and $p(b,a)$ denotes joint probability distribution.

	\subsection{Binning-based MI Estimator}
\begin{figure}[htbp]
		\centering
		
		\subfigure[sparse connectivity]{
			\includegraphics[width=0.3\textwidth]{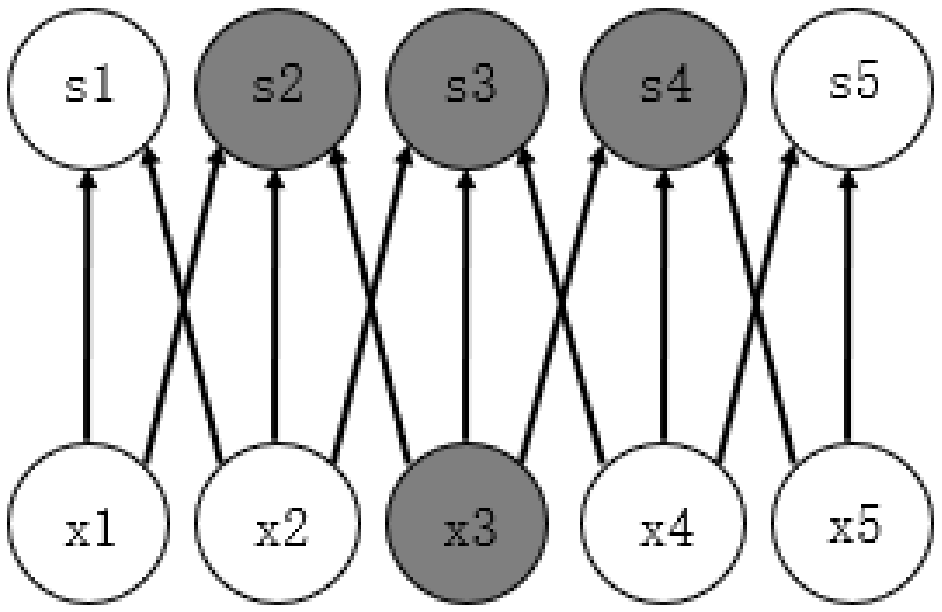}}
		\subfigure[dense connectivity ]{
			\includegraphics[width=0.3\textwidth]{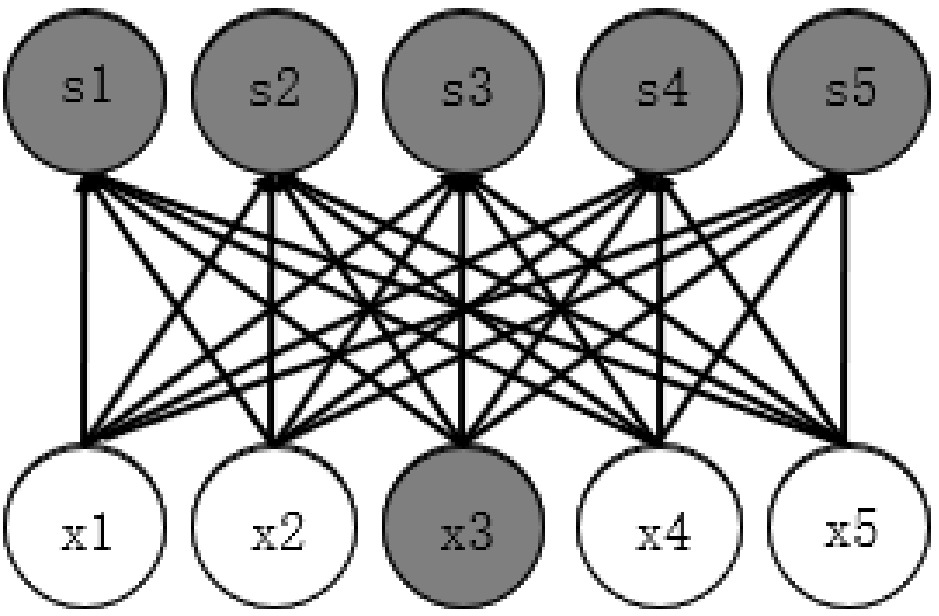}
		}
		\caption{Sparse connectivity and dense connectivity (fully connected) . }
		\label{sparse_dense}
	\end{figure}

\begin{figure}[htbp]
		\centering
		\includegraphics[width=0.7\textwidth]{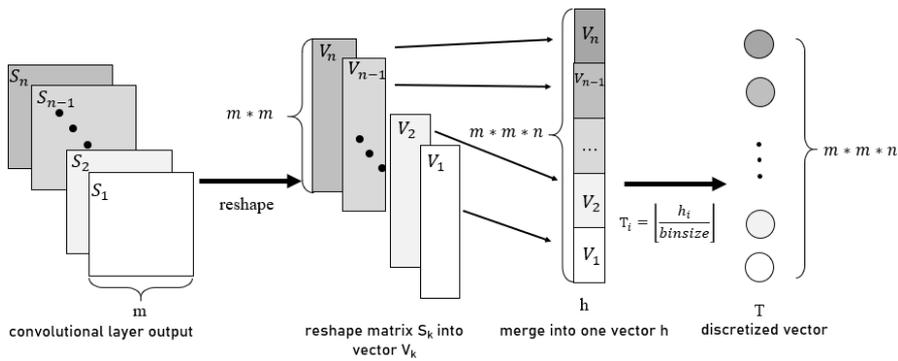}
		\caption{The process of binning activity. Each channel of layer's output is firstly converted into a vector, and then these vectors are combined into a long one, i.e. $h$. Then, it is mapped into a new vector by dividing by the constant bin size.}
		\label{binning}
	\end{figure}

	The binning-based MI estimator is widely used in feedforward  neural networks. And as we know, CNNs are characterized as sparse interactions (sparse connectivity) in compare with feedforward neural networks as shown in Fig.~\ref{sparse_dense}, which appears in Ref.~\citep{goodfellow2016deep}. In principle, the sparsity will not lead to the failure of binning-based estimator. So along this way, we also use it to evaluate the MI in CNNs. First, we reshape the output images of each channel of each convolutional layer into a vector, and splice these vectors into a long one $h$. Then, according to Ref.~\citep{saxe2019information}, we discretize the activation output by a fixed bin size, i.e. $T = bin(h)$. (Ref.~\citep{saxe2019information} choose 0.5, while in this paper, we use the constant 0.67 as bin size. Because according to our experimental results, it is good for visualization and meets the results of kernel density estimation method in~\citep{kolchinsky2017estimating, kolchinsky2019nonlinear}.) We show the process in Fig.~\ref{binning}.
	
	\begin{figure}[htbp]
		\centering
		\includegraphics[width=0.7\textwidth]{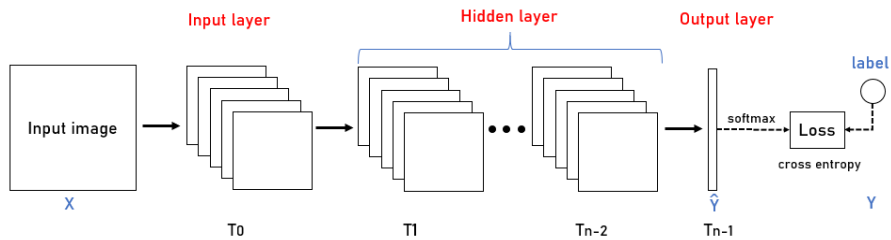}
		\caption{Architecture of our convolutional network model. $X$ denotes the input data and $Y$ is the label.} $T_0$ is input layer. $T_0 \sim T_{n-2}$ are convolutional layers. $T_{n-1}$ is fully connected output layer. 
		\label{network}
	\end{figure}
	
			\begin{figure*}[htbp]
		\centering
		\subfigure[MI path on training data of MNIST]{
			\includegraphics[width=0.8\textwidth]{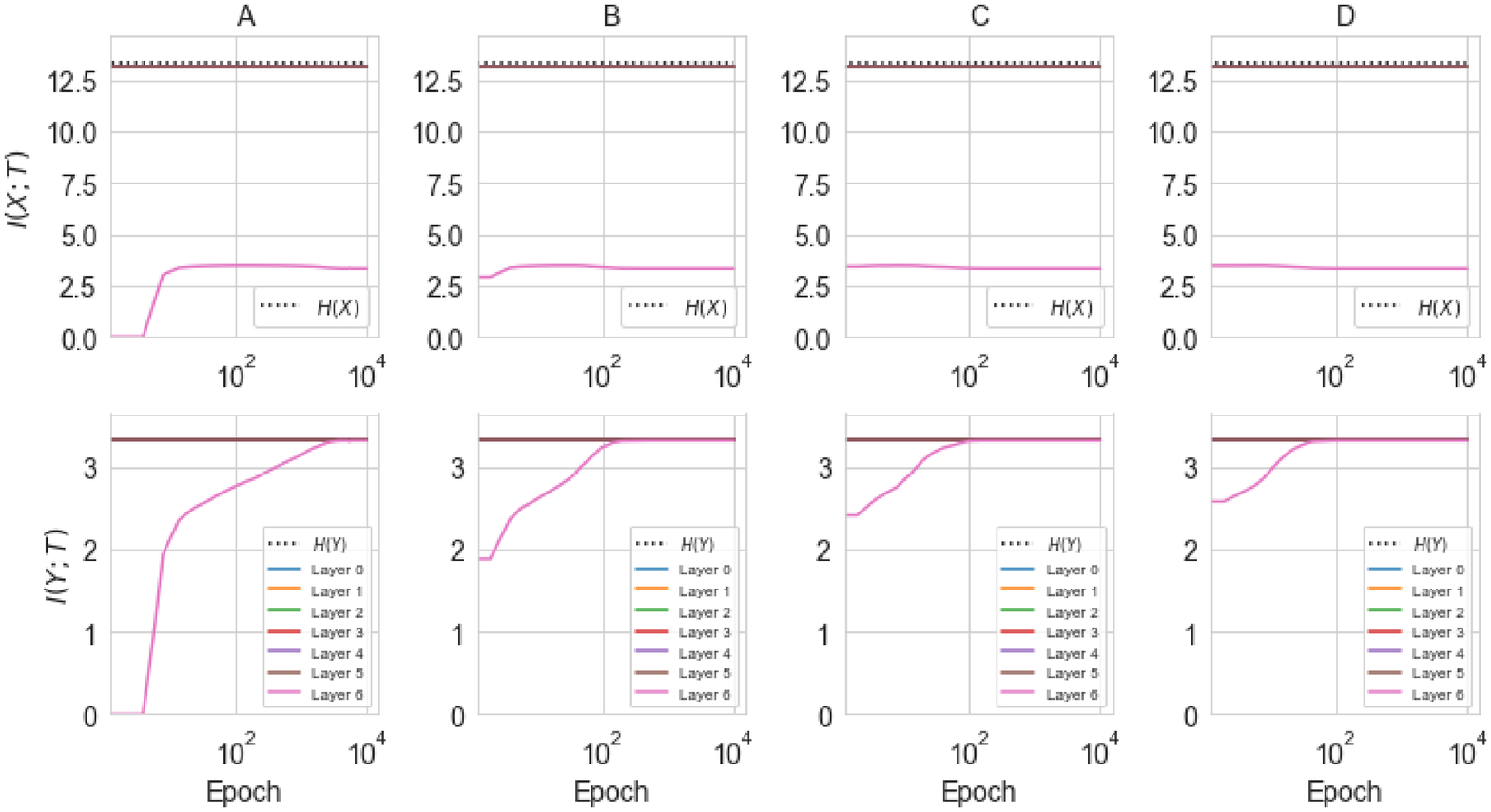}}
		\subfigure[MI path on training data of Fashion-MNIST]{
			\includegraphics[width=0.8\textwidth]{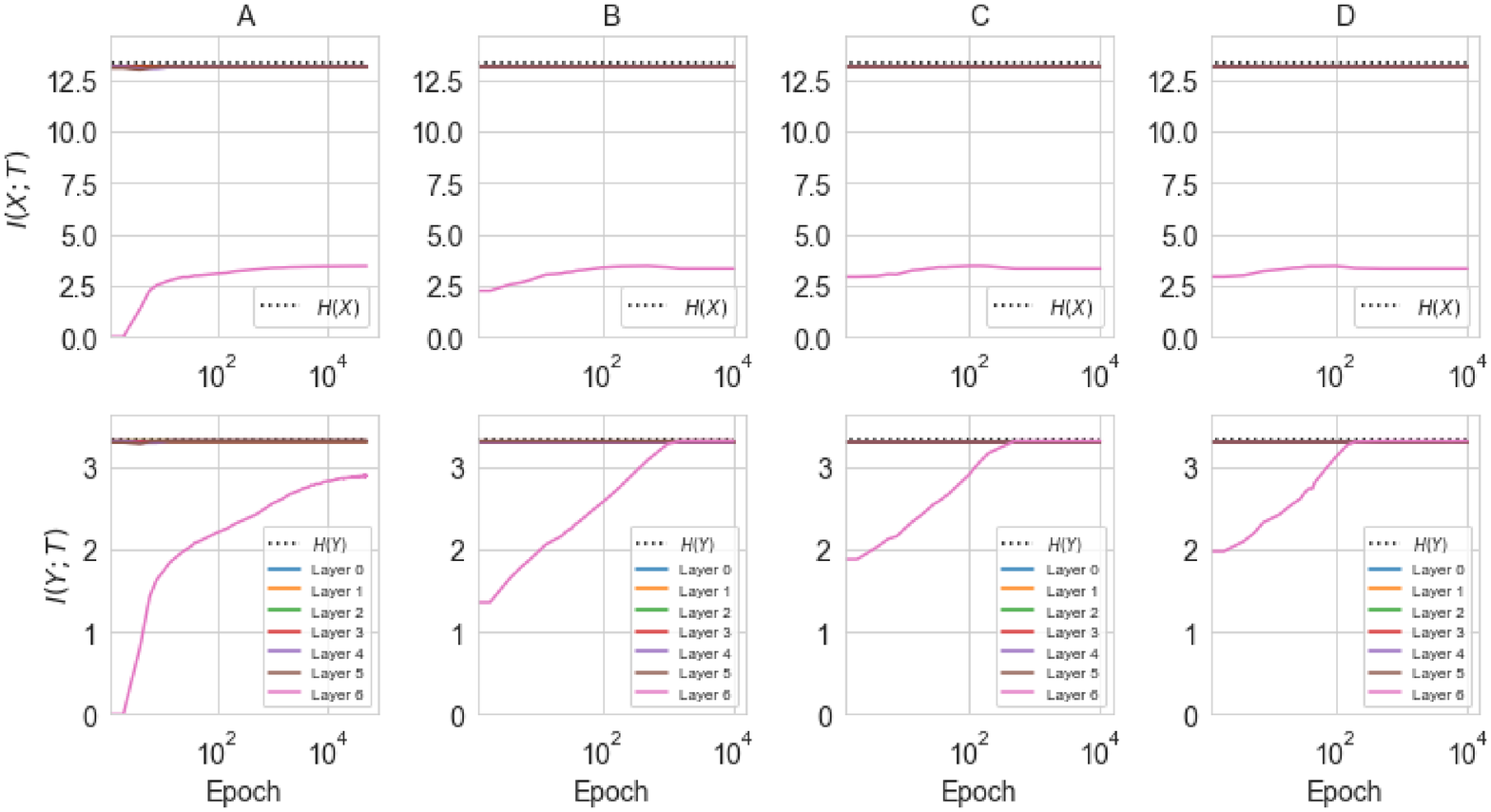}
		}
		\subfigure[MI path on test data of MNIST]{
			\includegraphics[width=0.8\textwidth]{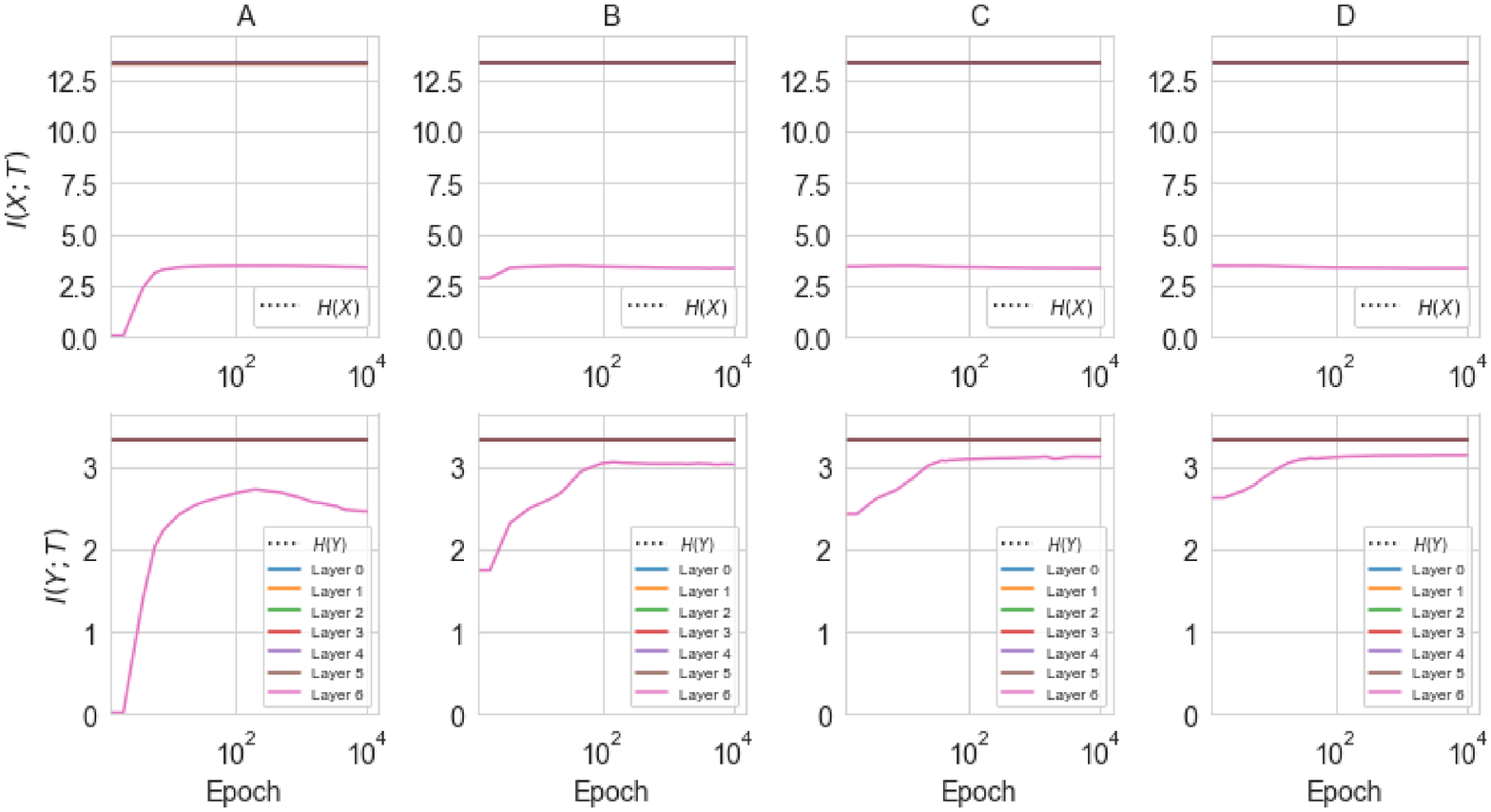}
		}
		\end{figure*}
		\addtocounter{figure}{-1}
		\begin{figure*}[htbp]
		\centering
		\addtocounter{figure}{1}
		\subfigure[MI path on test data of Fashion-MNIST]{
			\includegraphics[width=0.8\textwidth]{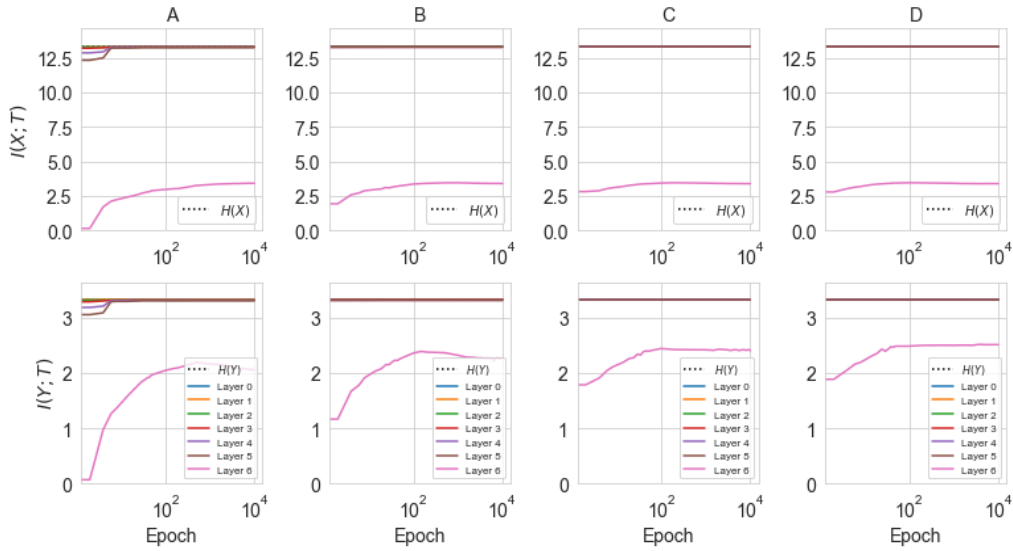}
		}
		\caption{(Colored online) MI path on CNNs with different convolutional layer widths. Colored lines mark each layer of network. As we describe in \textit{Experiments and discussion} section, layer $0 \sim 5$ are convolutional layers, and layer 6 is final output layer (fully connected layer). In this experiment, layer 0 matches layer $T_0$ in Fig.~\ref{network}, layer 5 matches layer $T_{n-2}$ in Fig~\ref{network} and layer 6 matches layer $T_{n-1}$ in Fig.~\ref{network}.} 
			Convolutional layer 5 (the final convolutional layer) covers all previous layers, $H(X)$ and $H(Y)$ since they all have the same value. The convolutional layer width of 4 networks are (A) 1-1-1-1-1-1, (B) 3-3-3-3-3-3, (C) 6-6-6-6-6-6, (D) 12-12-12-12-12-12. The pink line represents the mutual information of final output layer, which grows until getting stable as the process of training. The kernel size of these networks are set to 3x3.
		\label{widthMI}
	\end{figure*}

	\begin{figure*}[htbp]
		\centering
		
		\subfigure[MI path on training data of MNIST. All networks with fixed depth = 3. The sizes of kernel are (A) 3 x 3 (B) 7 x 7 (C) 11 x 11]
		{\includegraphics[width=0.65\textwidth]{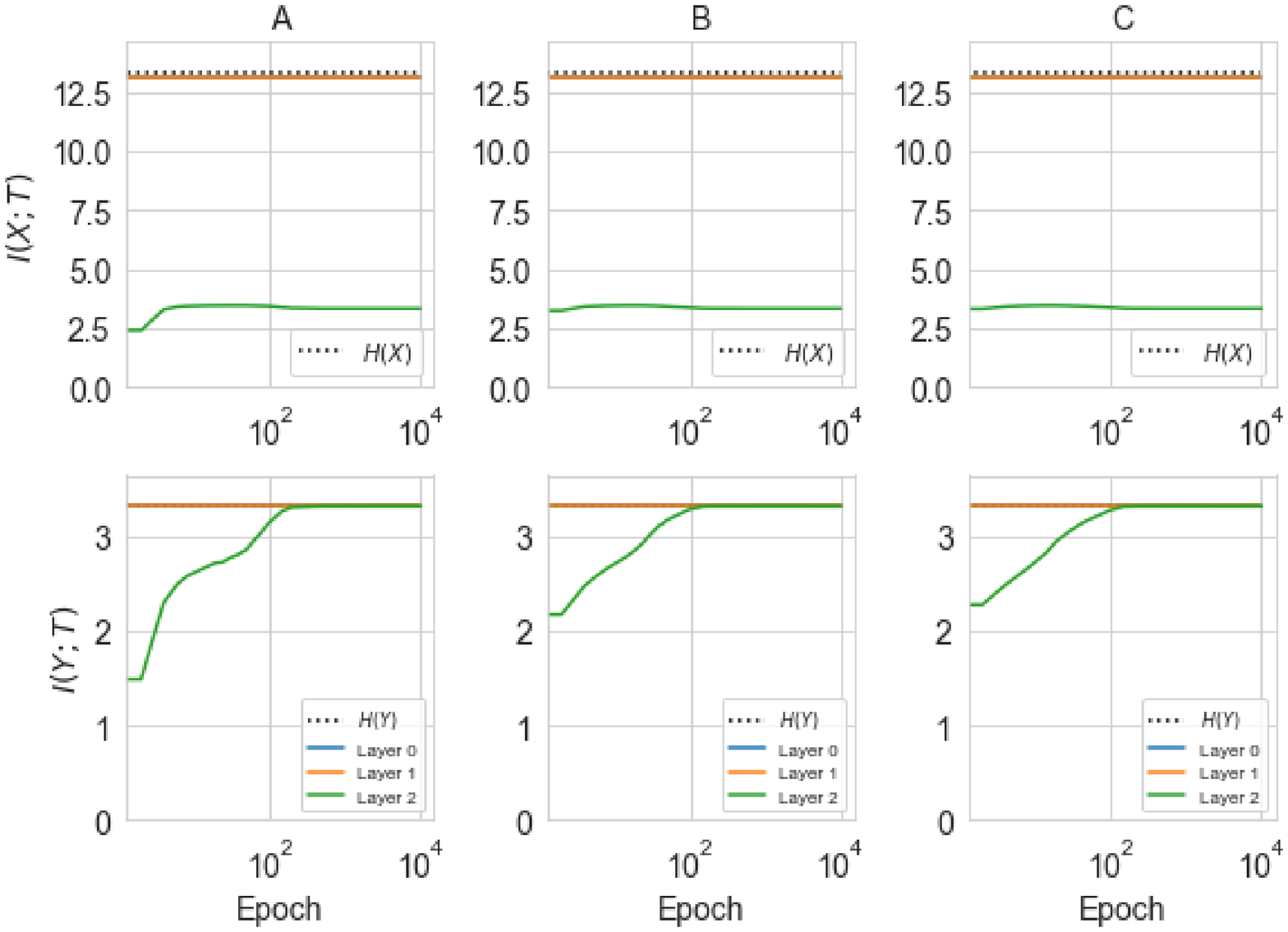}}
		\subfigure[ MI path on training data of MNIST. All networks with fixed depth = 6. The sizes of kernel are (A) 3 x 3 (B) 5 x 5 (C) 7 x 7]{
		\includegraphics[width=0.65\textwidth]{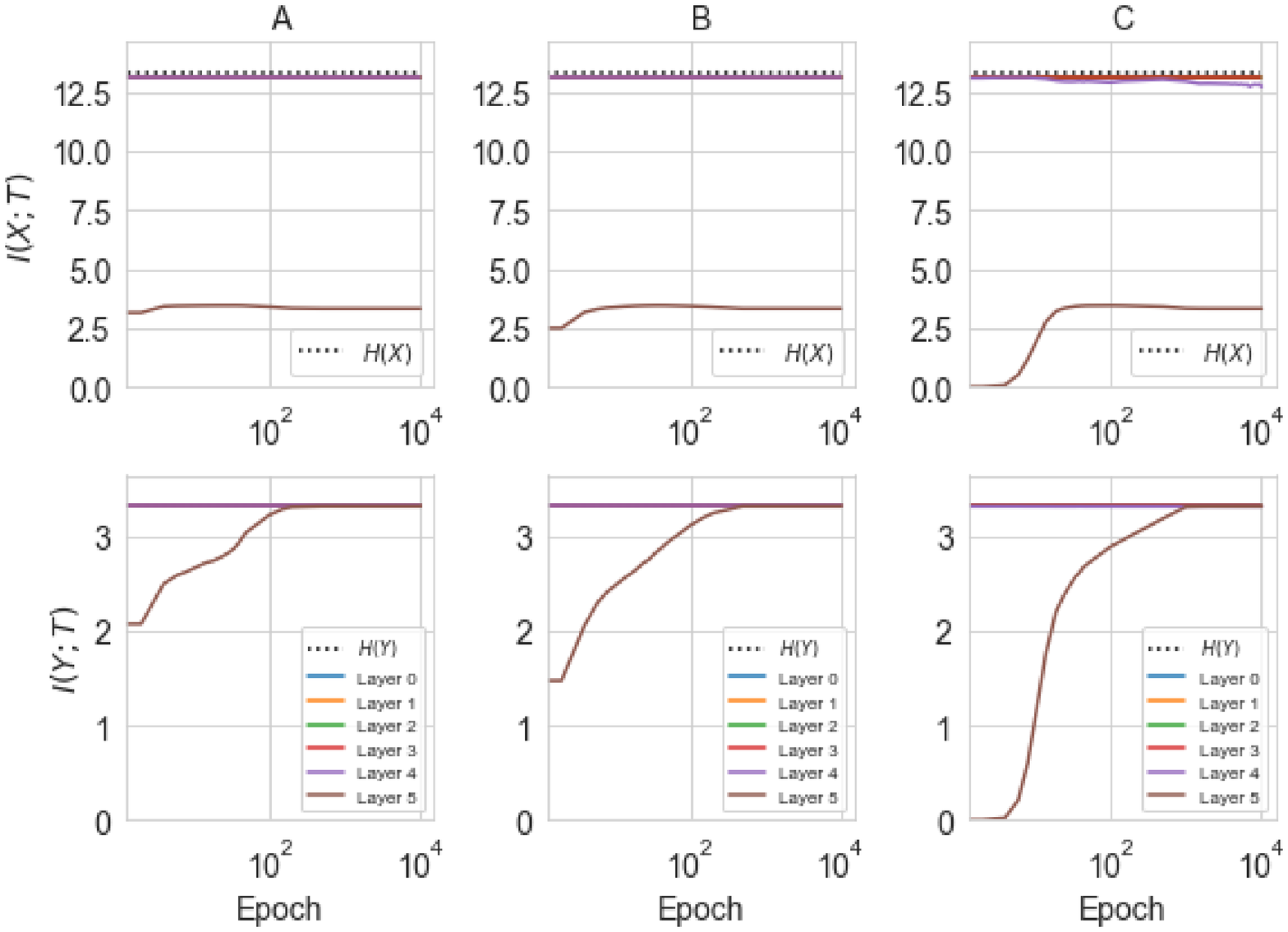}}
		
		\subfigure[MI path on training data of Fashion-MNIST. All networks with fixed depth = 3. The sizes of kernel are (A) 3 x 3 (B) 7 x 7 (C) 11 x 11]{
			\includegraphics[width=0.65\textwidth]{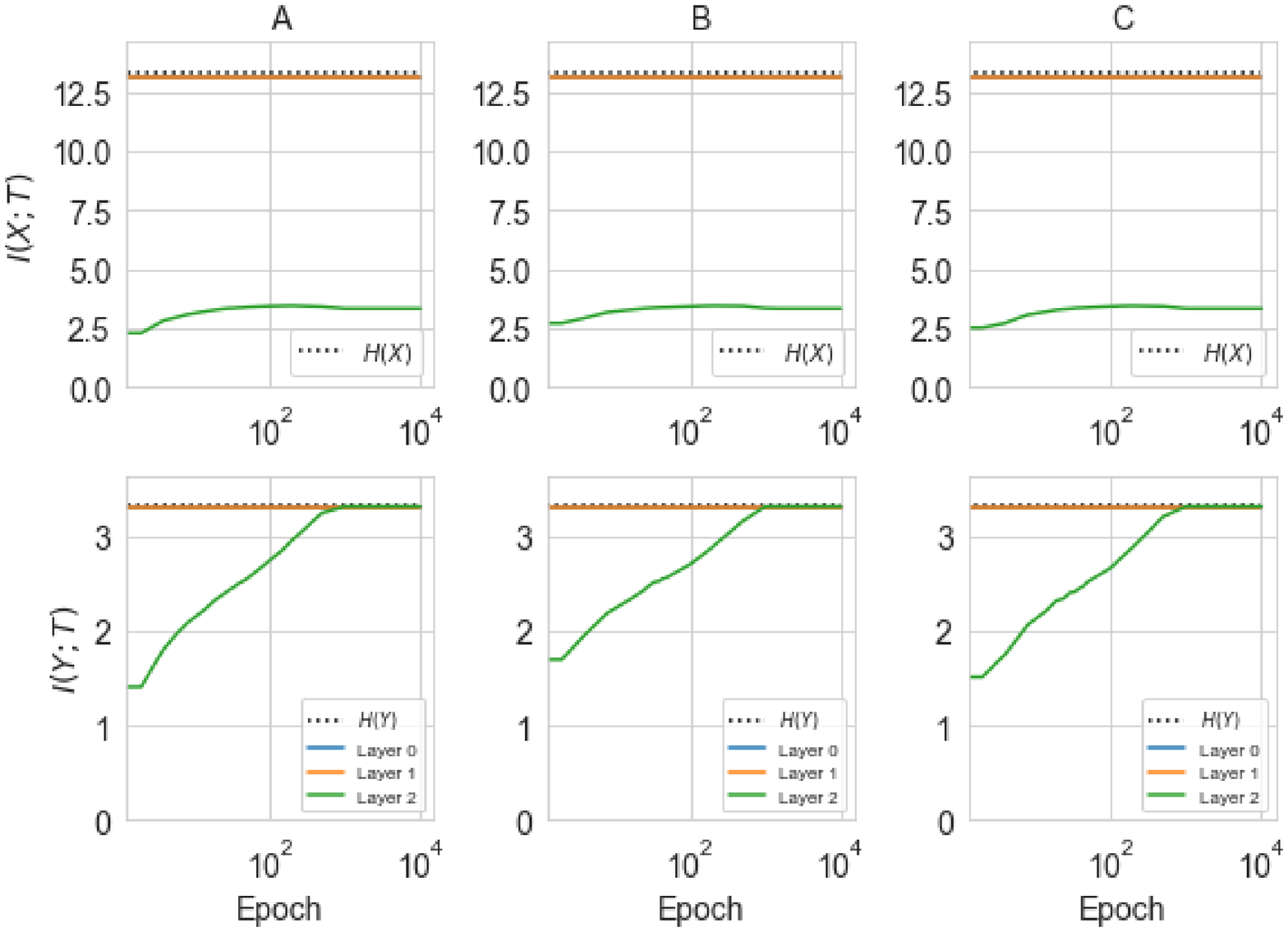}}
	\end{figure*}	
		\addtocounter{figure}{-1} 
		
	\begin{figure*}[htbp]
	\centering
		\addtocounter{figure}{-1} 
		\subfigure[~MI path on training data of Fashion-MNIST. All networks with fixed depth  = 6. The sizes of kernel are (A) 3 x 3 (B) 5 x 5 (C) 7 x 7]{
			\includegraphics[width=0.65\textwidth]{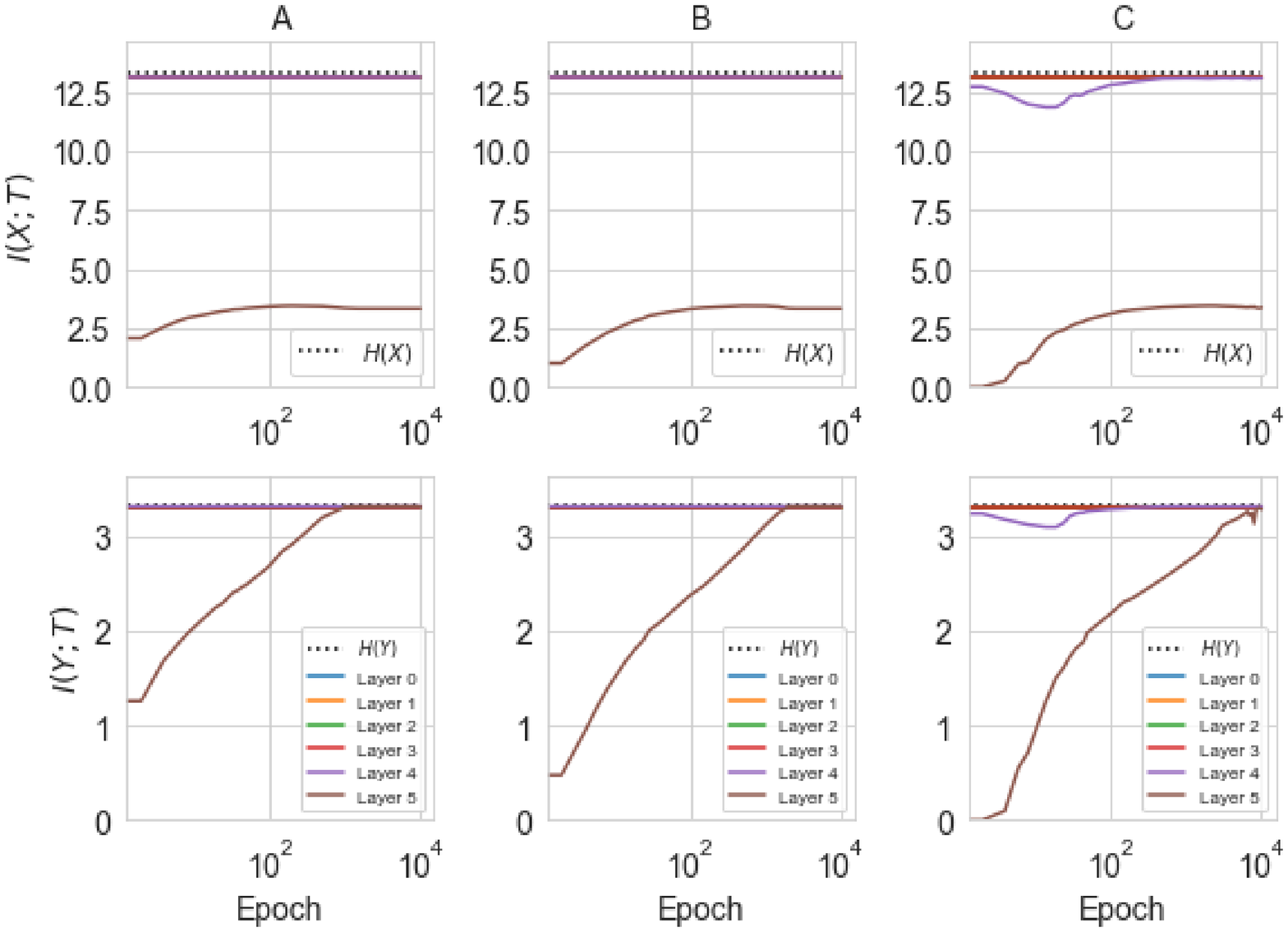}}
		
		\subfigure[MI path on test data of MNIST. All networks with fixed depth  = 3. The sizes of kernel are (A) 3 x 3 (B) 7 x 7 (C) 11 x 11]{
			\includegraphics[width=0.65\textwidth]{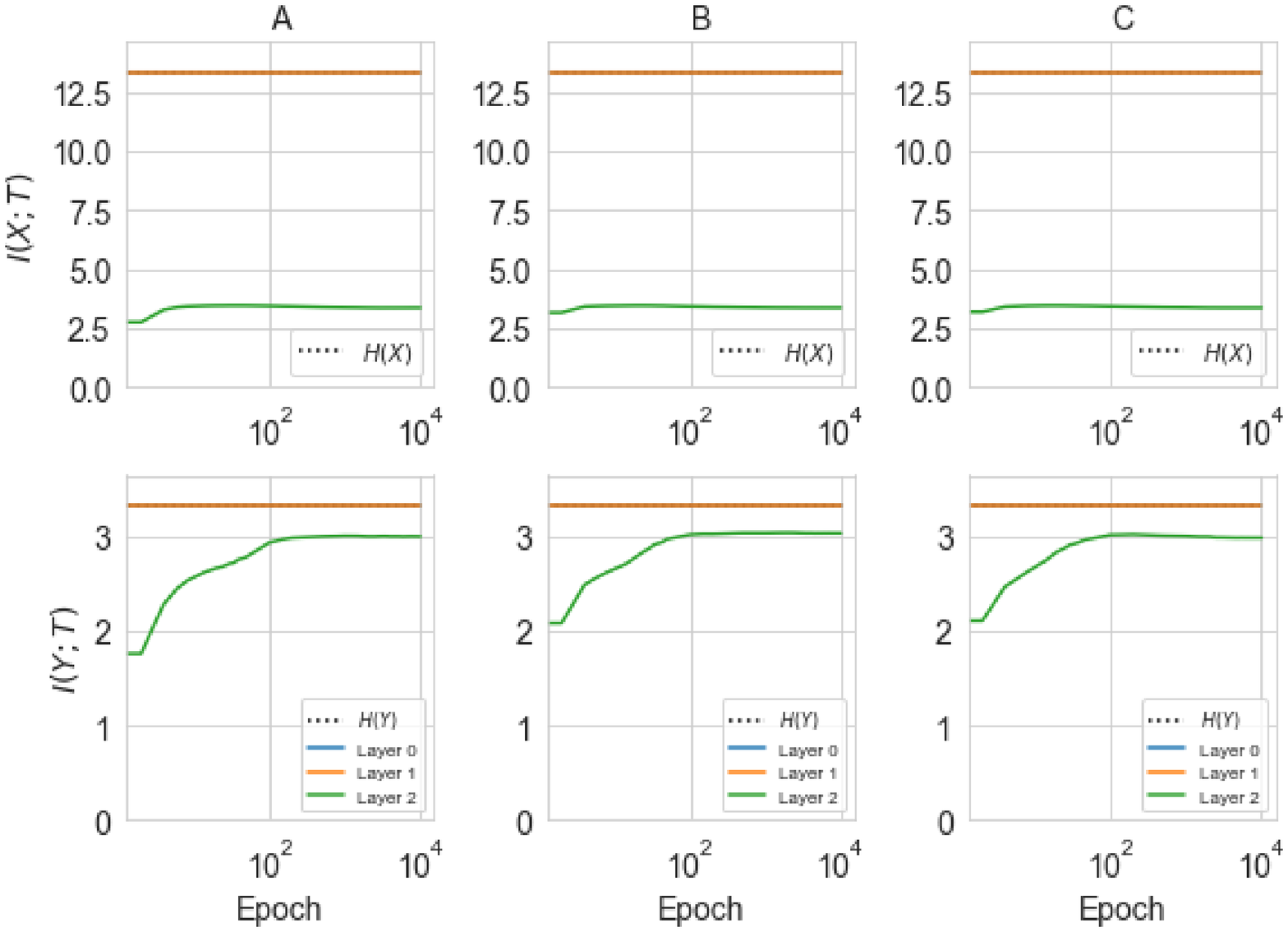}}
		\subfigure[ MI path on test data of MNIST. All networks with fixed depth size = 6. The sizes of kernel are (A) 3 x 3 (B) 5 x 5 (C) 7 x 7]{
			\includegraphics[width=0.65\textwidth]{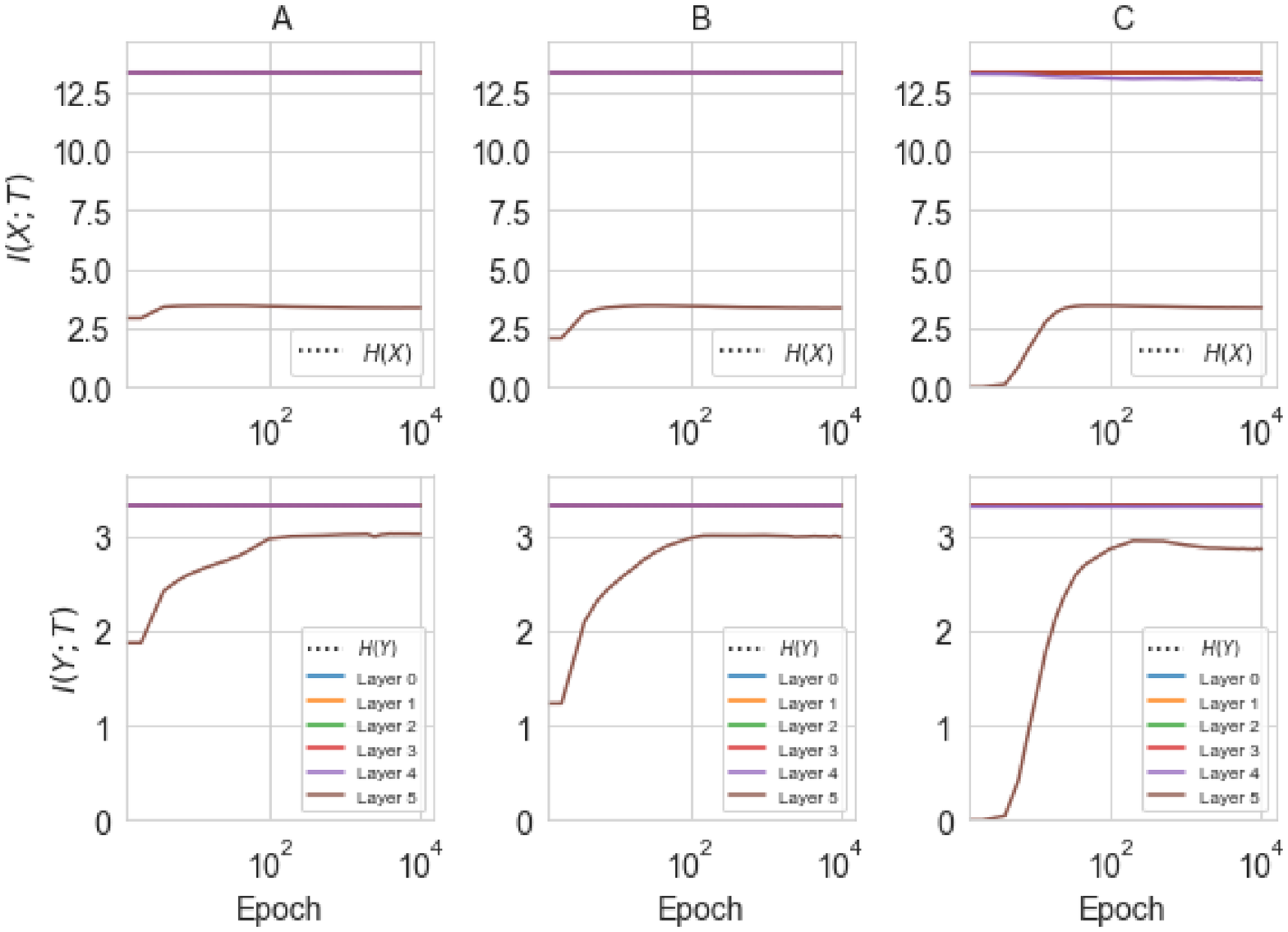}}
	\end{figure*}
		\addtocounter{figure}{1} 
		
		\begin{figure*}
		\centering
		\addtocounter{figure}{1} 
		\subfigure[ MI path on test data of Fashion-MNIST. All networks with fixed depth = 3. The sizes of kernel are (A) 3 x 3 (B) 7 x 7 (C) 11 x 11]{
			\includegraphics[width=0.65\textwidth]{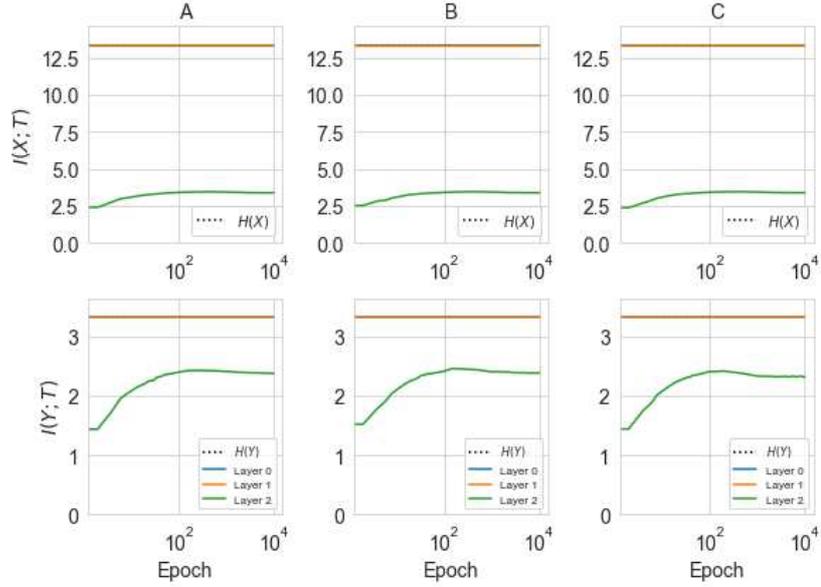}}
		\subfigure[ MI path on test data of Fashion-MNIST. All networks with fixed depth = 6. The sizes of kernel are (A) 3 x 3 (B) 5 x 5 (C) 7 x 7]{
			\includegraphics[width=0.65\textwidth]{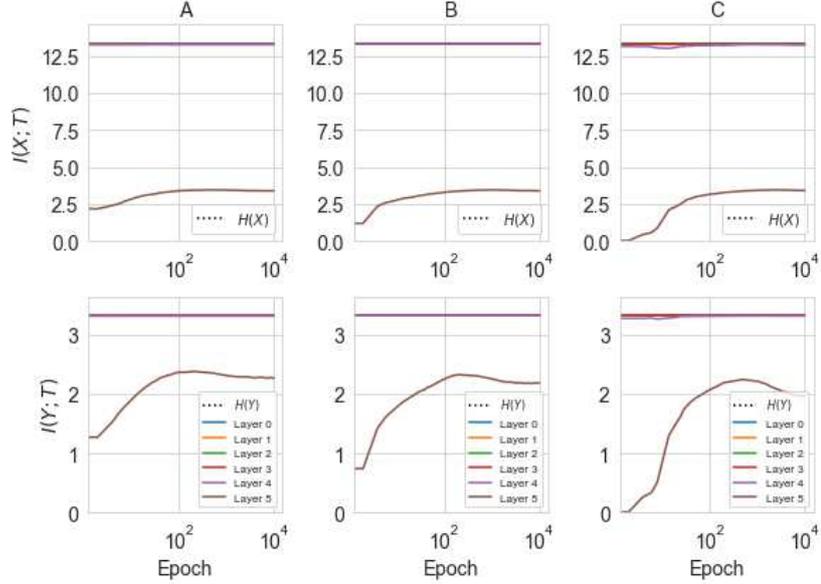}}

		\caption{(Colored online) MI path on CNNs with different convolutional kernel sizes. In this configuration, all convolutional layers width are set to 3.  
		}
		\label{Kernel}
	\end{figure*}
	
	\begin{figure*}[htbp]
		\centering
		\subfigure[MI path on training data of MNIST. (A) depth=2, (B) depth=3, (C) depth=7, (D) depth=10]{
			\includegraphics[width=0.9\textwidth]{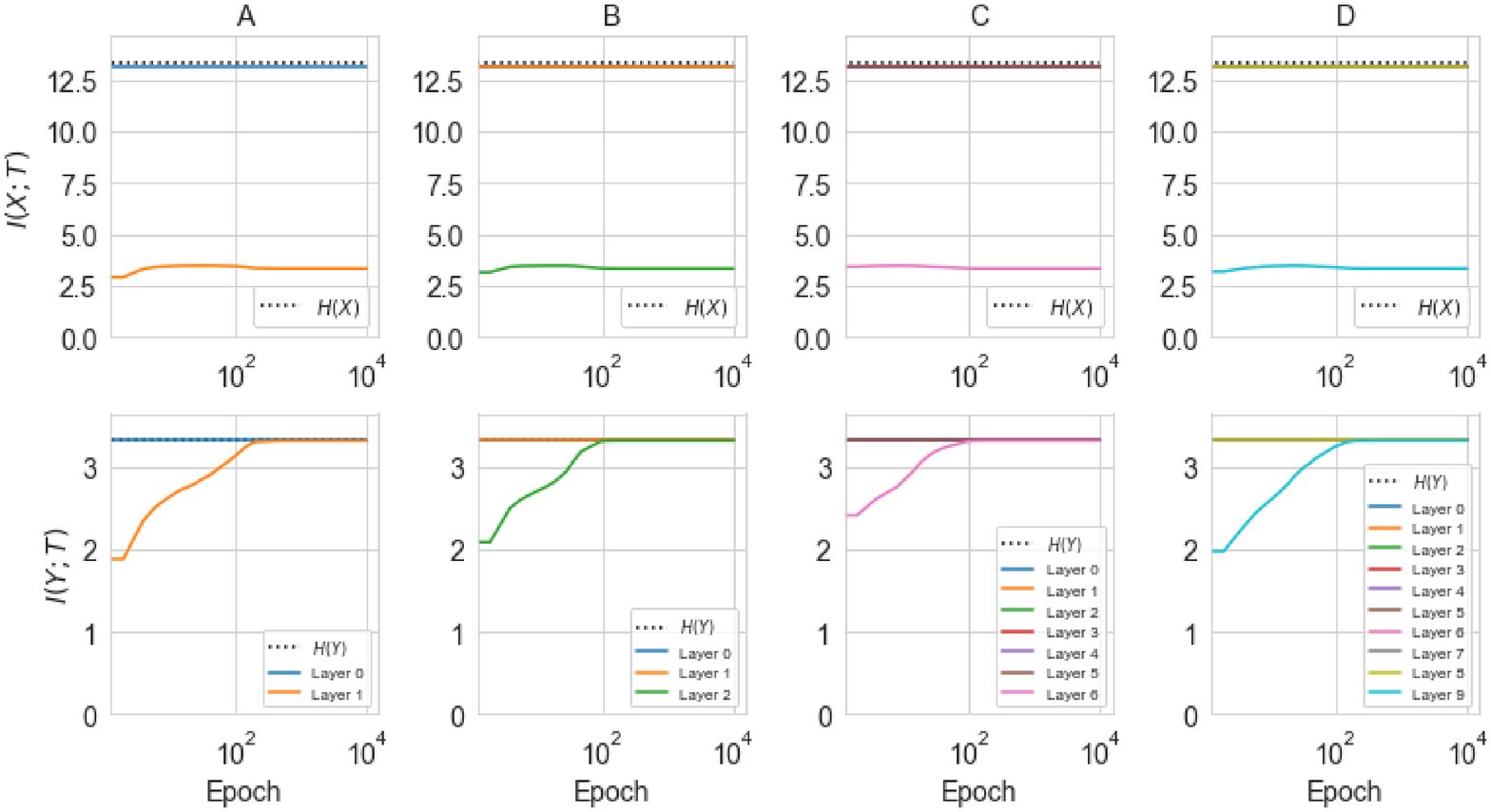}}
		\subfigure[MI path on training data of Fashion-MNIST. (A) depth=2, (B) depth=3, (C) depth=7, (D) depth=10]{
			\includegraphics[width=0.9\textwidth]{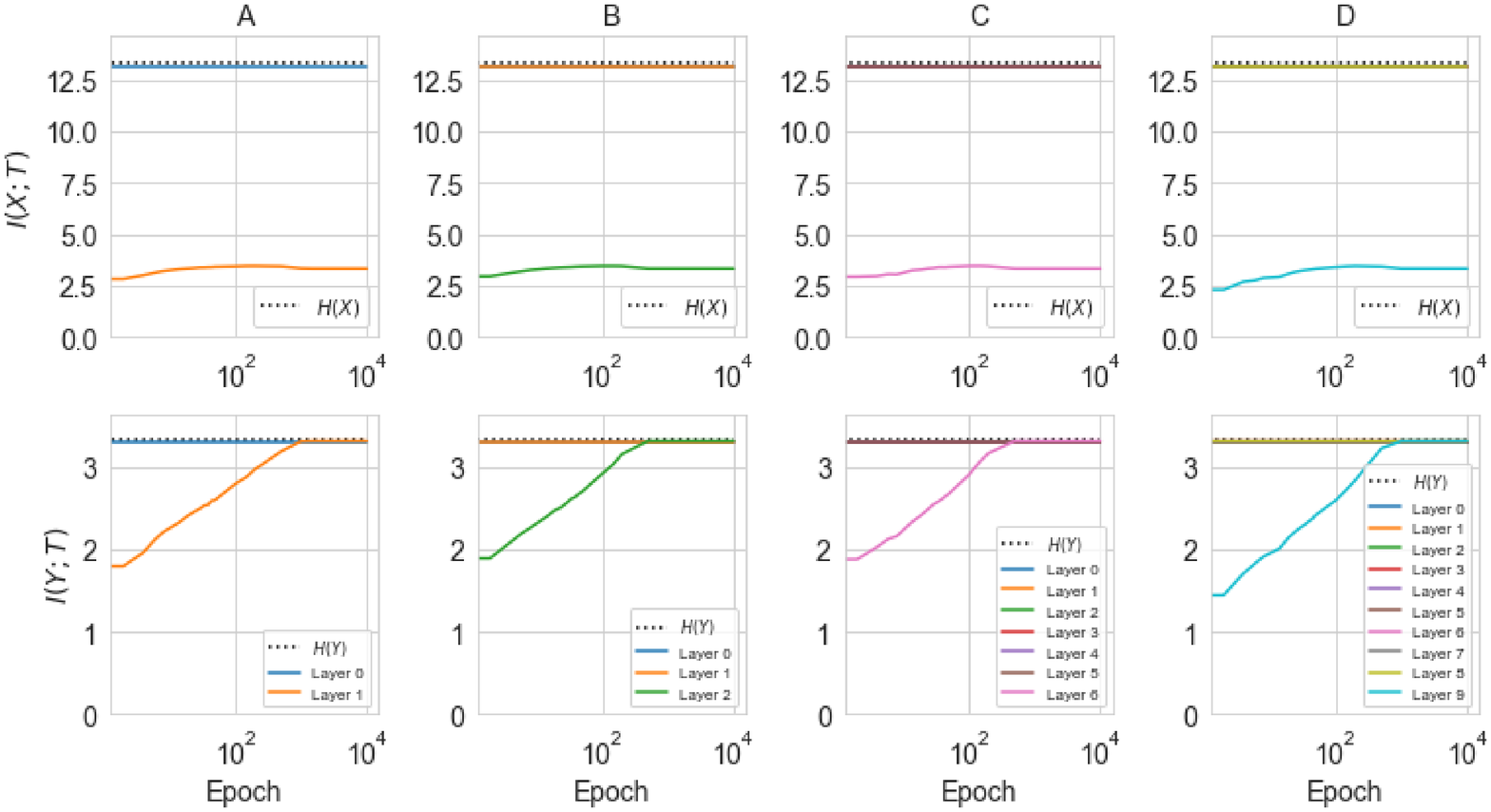}}
		\subfigure[MI path on test data of MNIST. (A) depth=2, (B) depth=3, (C) depth=7, (D) depth=10]{
			\includegraphics[width=0.9\textwidth]{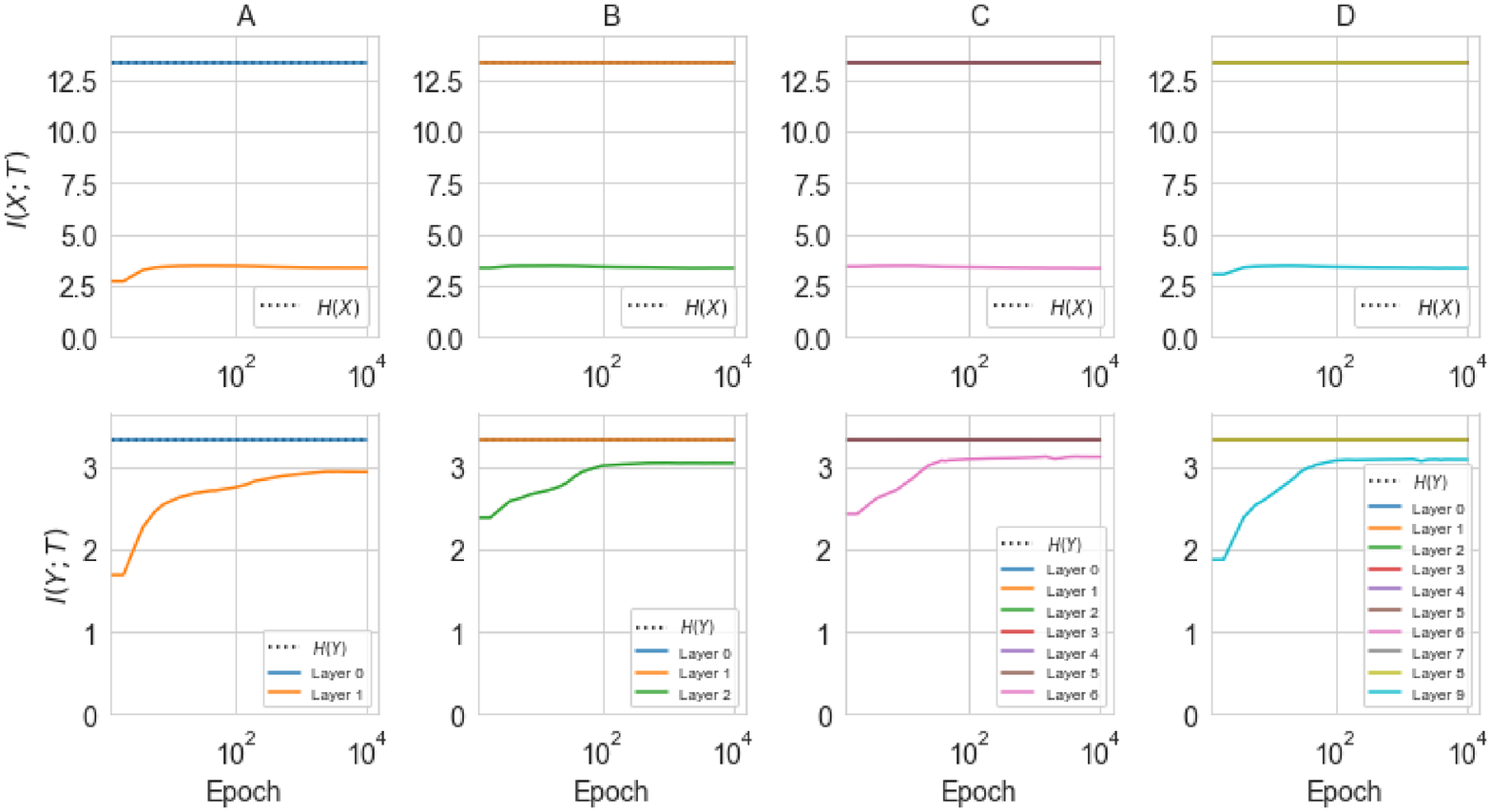}}
					\end{figure*}	
		\addtocounter{figure}{-1} 
		
		\begin{figure*}
			\addtocounter{figure}{1} 
		\centering
		\subfigure[MI path on test data of Fashion-MNIST. (A) depth=2, (B) depth=3, (C) depth=7, (D) depth=10]{
			\includegraphics[width=0.9\textwidth]{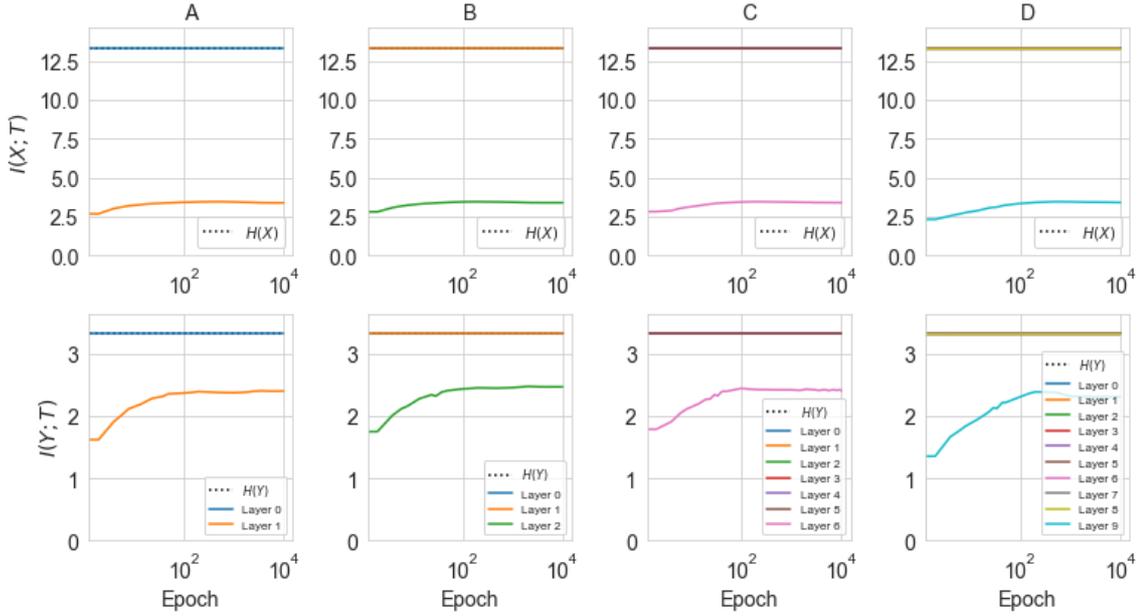}}
		
		\caption{(Colored online) MI path on CNNs with different depths. All width of convolutional layers are 6, and all kernel sizes are set to 3x3. } 
		\label{Depth}
	\end{figure*}
	
	\begin{figure*}[htbp]
		\centering
		\subfigure[MI on training data of MNIST. (A) without pooling layer (B) with pooling layer ]{
			\includegraphics[width=0.45\textwidth]{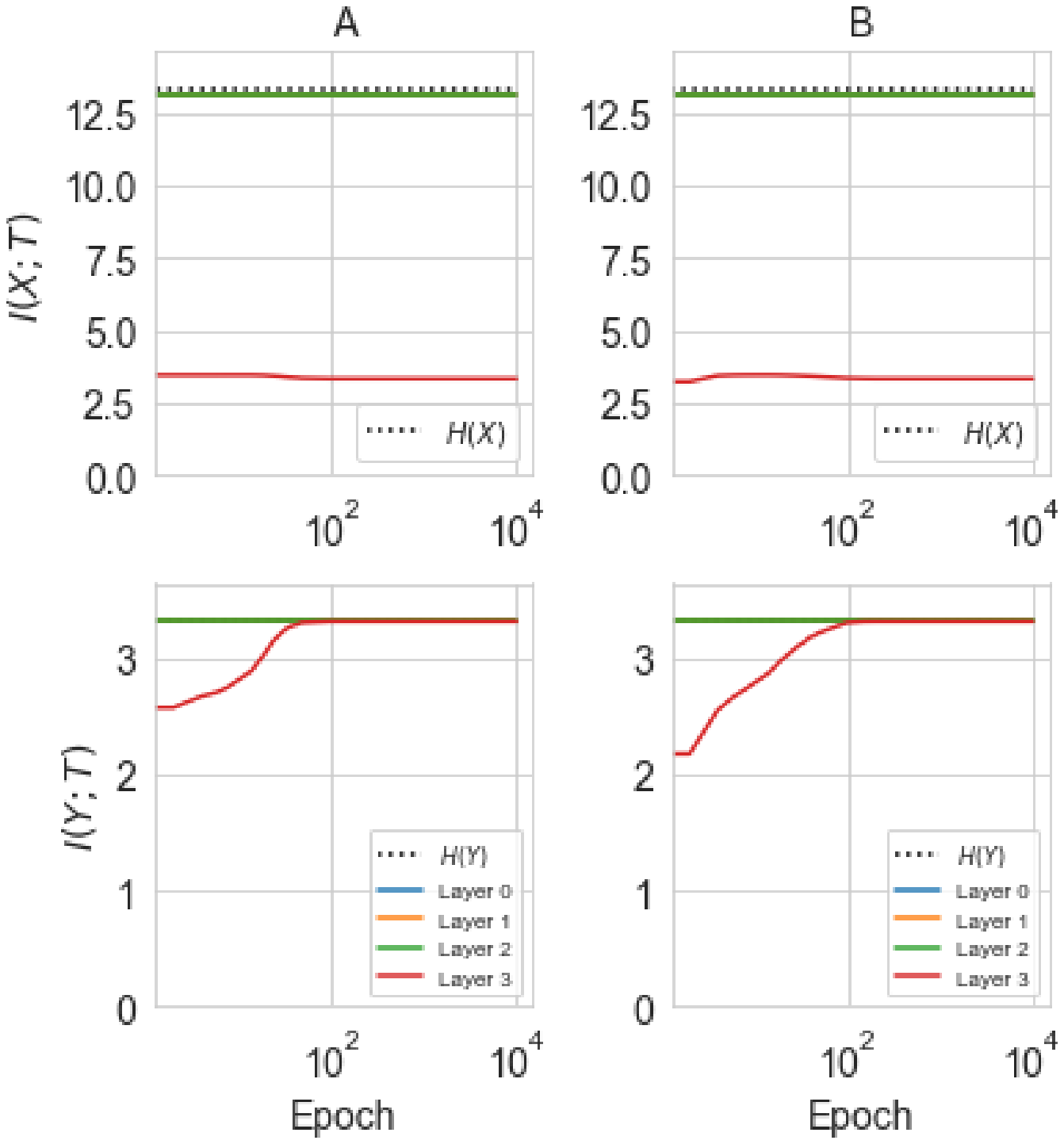}}
		\subfigure[MI on training data of Fashion-MNIST. (A) without pooling layer (B) with pooling layer]{
			\includegraphics[width=0.45\textwidth]{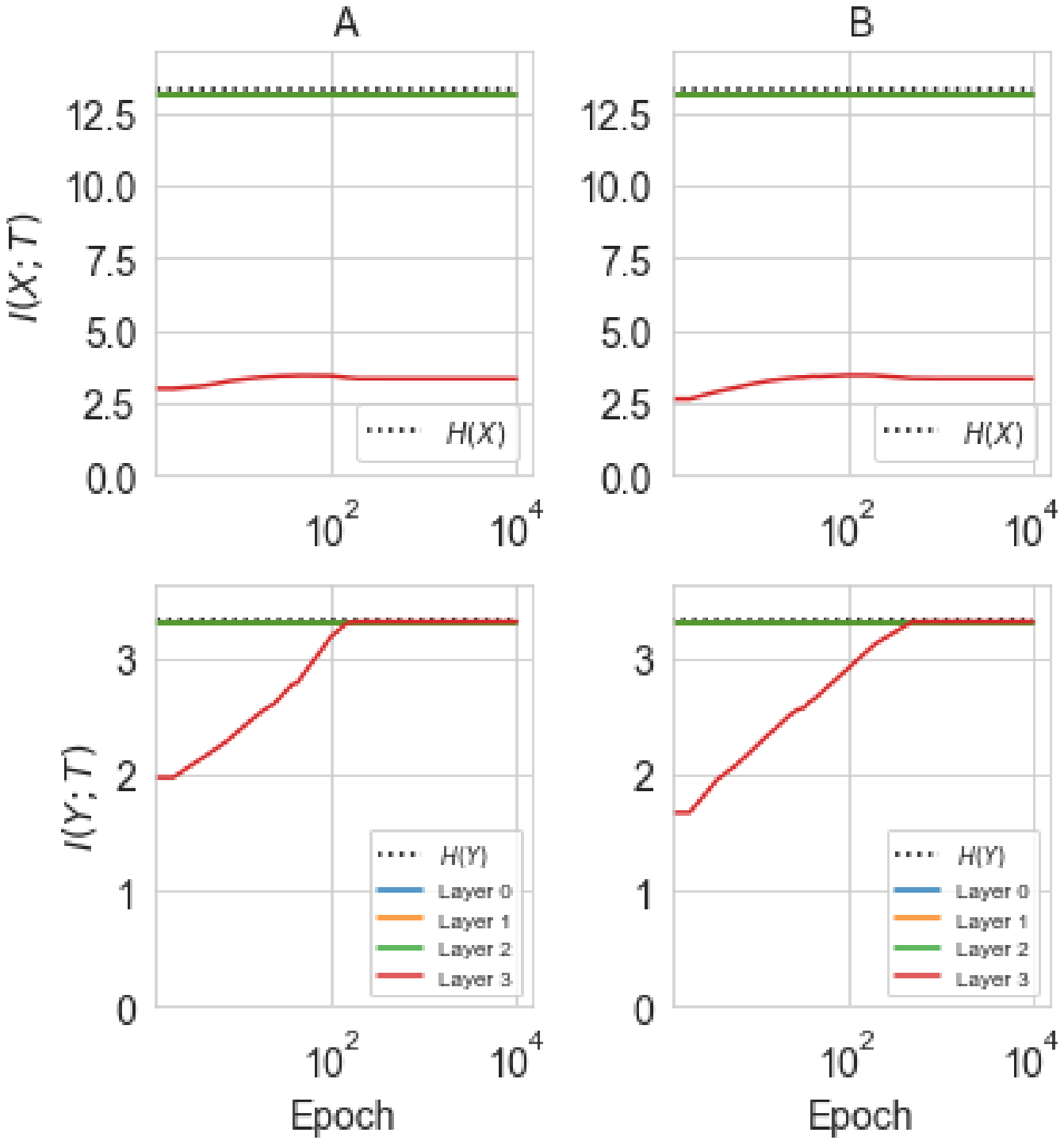}}
			
		\subfigure[MI on test data of MNIST. (A) without pooling layer (B) with pooling layer]{
			\includegraphics[width=0.45\textwidth]{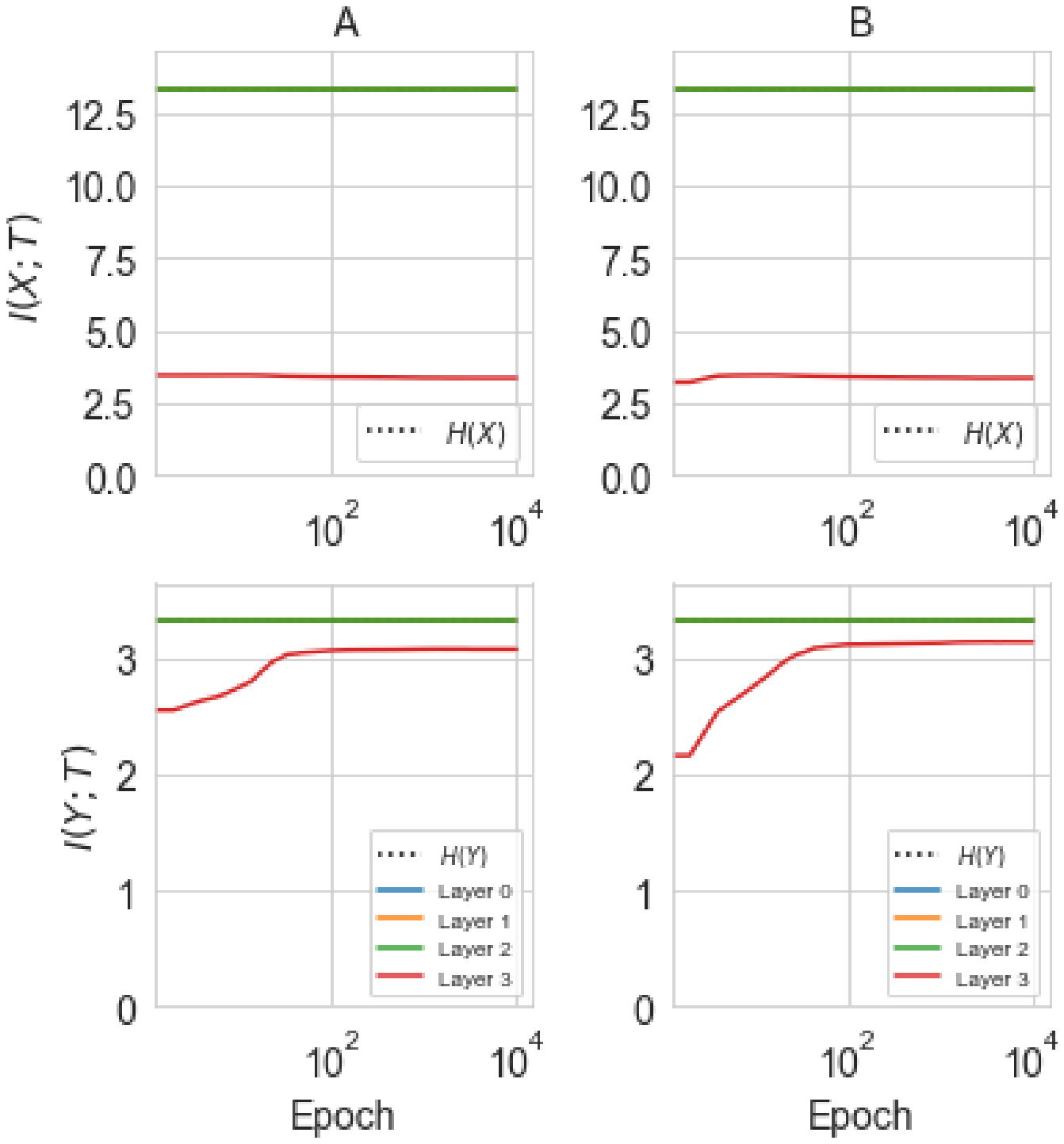}}
		\subfigure[MI on test data of Fashion-MNIST. (A) without pooling layer (B) with pooling layer]{
			\includegraphics[width=0.45\textwidth]{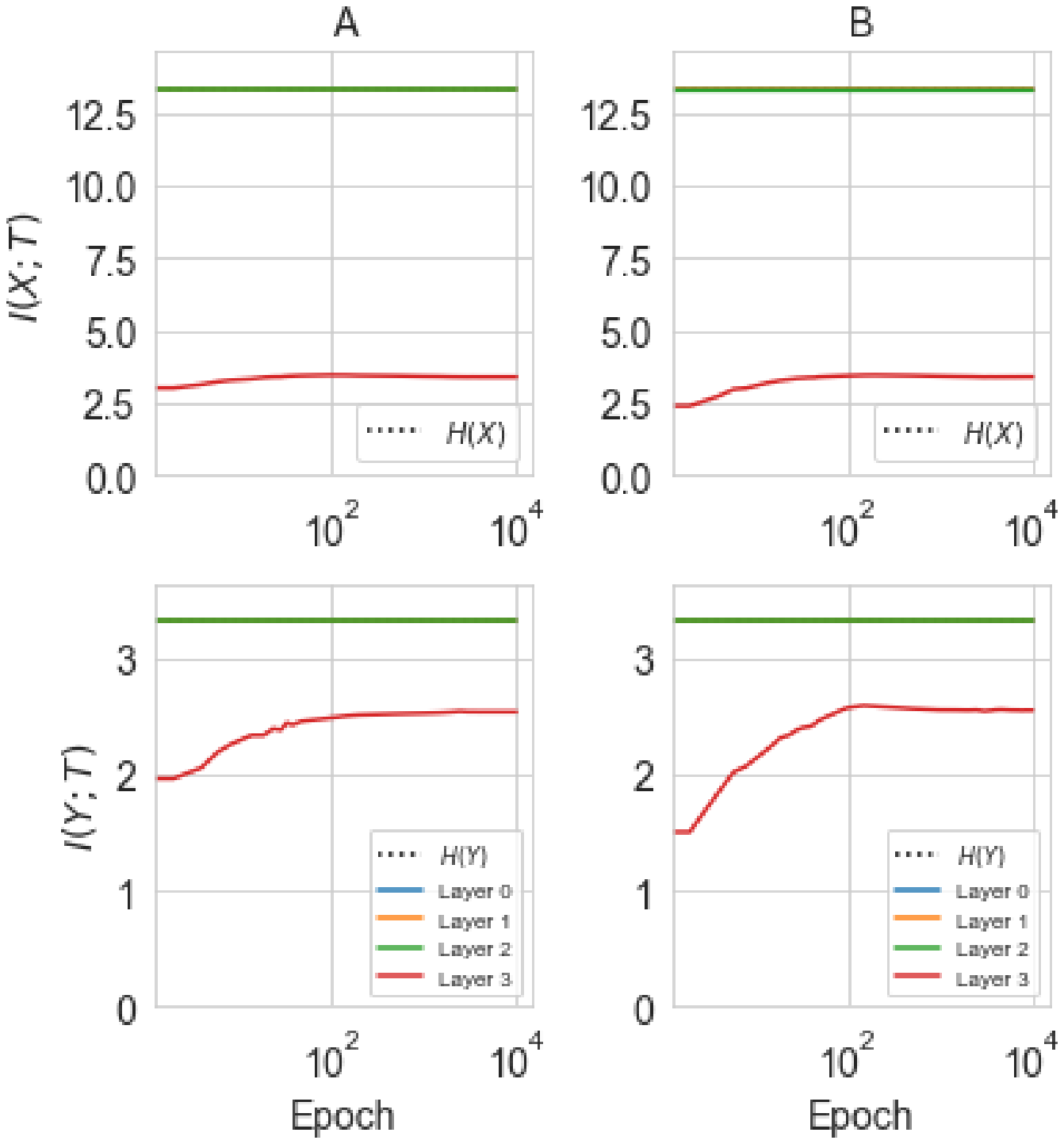}}
		
		\caption{(Colored online) MI path on CNNs with pooling layer. The width of all convolutional layers are 12-12-12 and the kernel size is fixed to 3x3. Moreover, the MaxPooling2D layer is added after the layer 1 and the pooling size is 2x2. }
		\label{pool}
	\end{figure*}

		\begin{figure*}[htbp]
		\centering
		\subfigure[IP on training data of MNIST]{
			\includegraphics[width=0.35\textwidth]{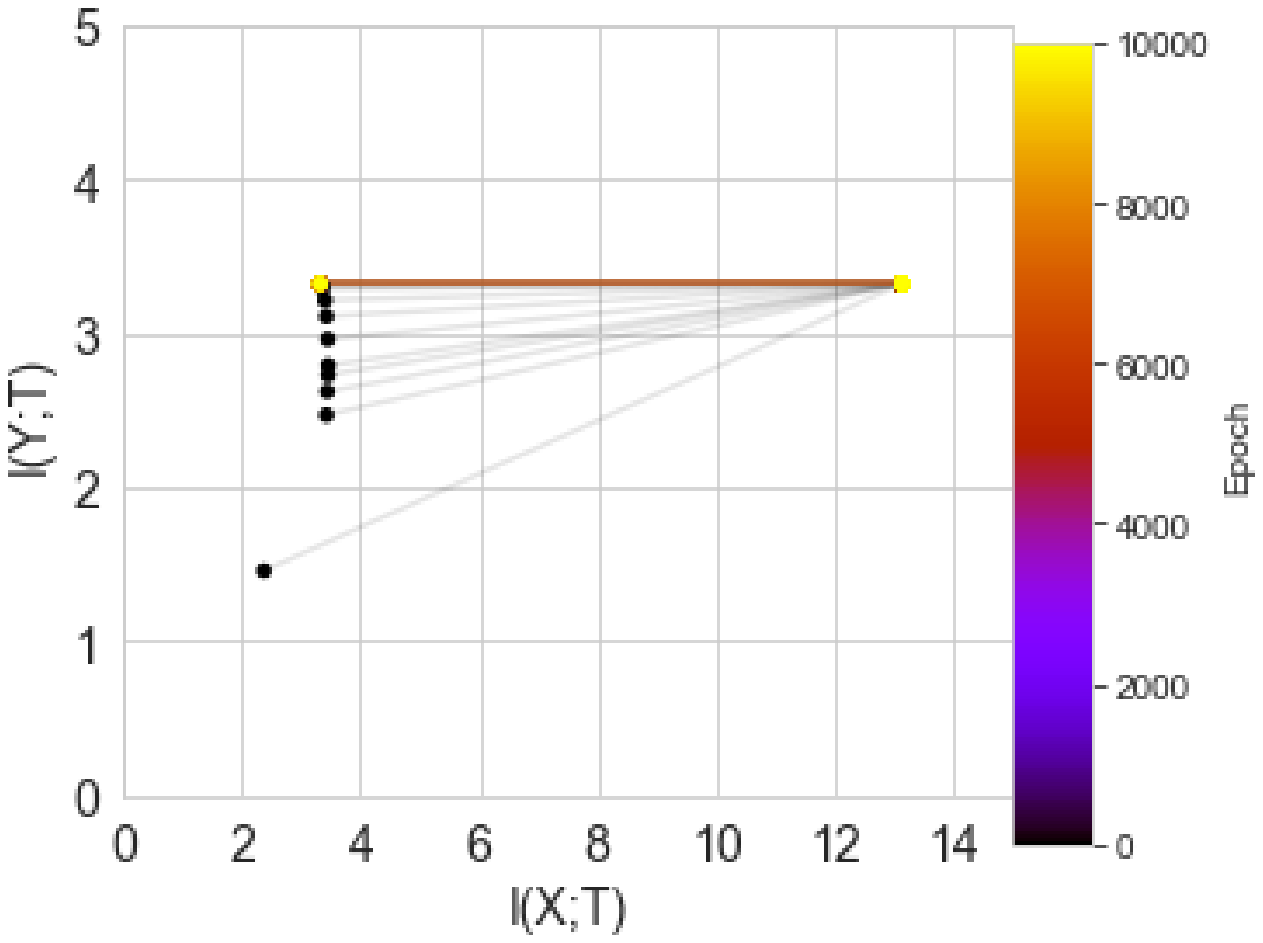}}
		\subfigure[IP on training data of Fashion-MNIST ]{
			\includegraphics[width=0.35\textwidth]{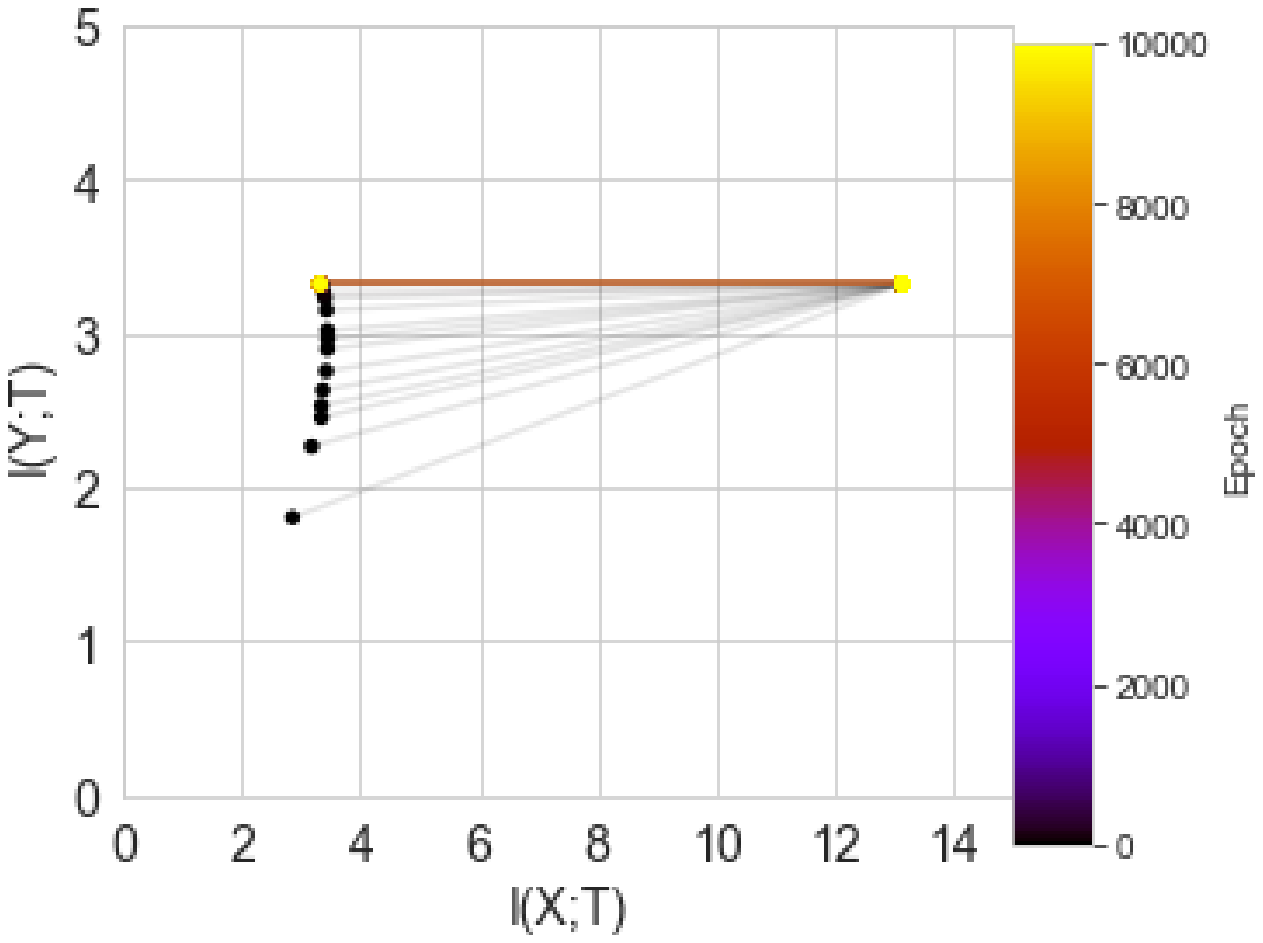}}
		\subfigure[IP on test data of MNIST ]{
			\includegraphics[width=0.35\textwidth]{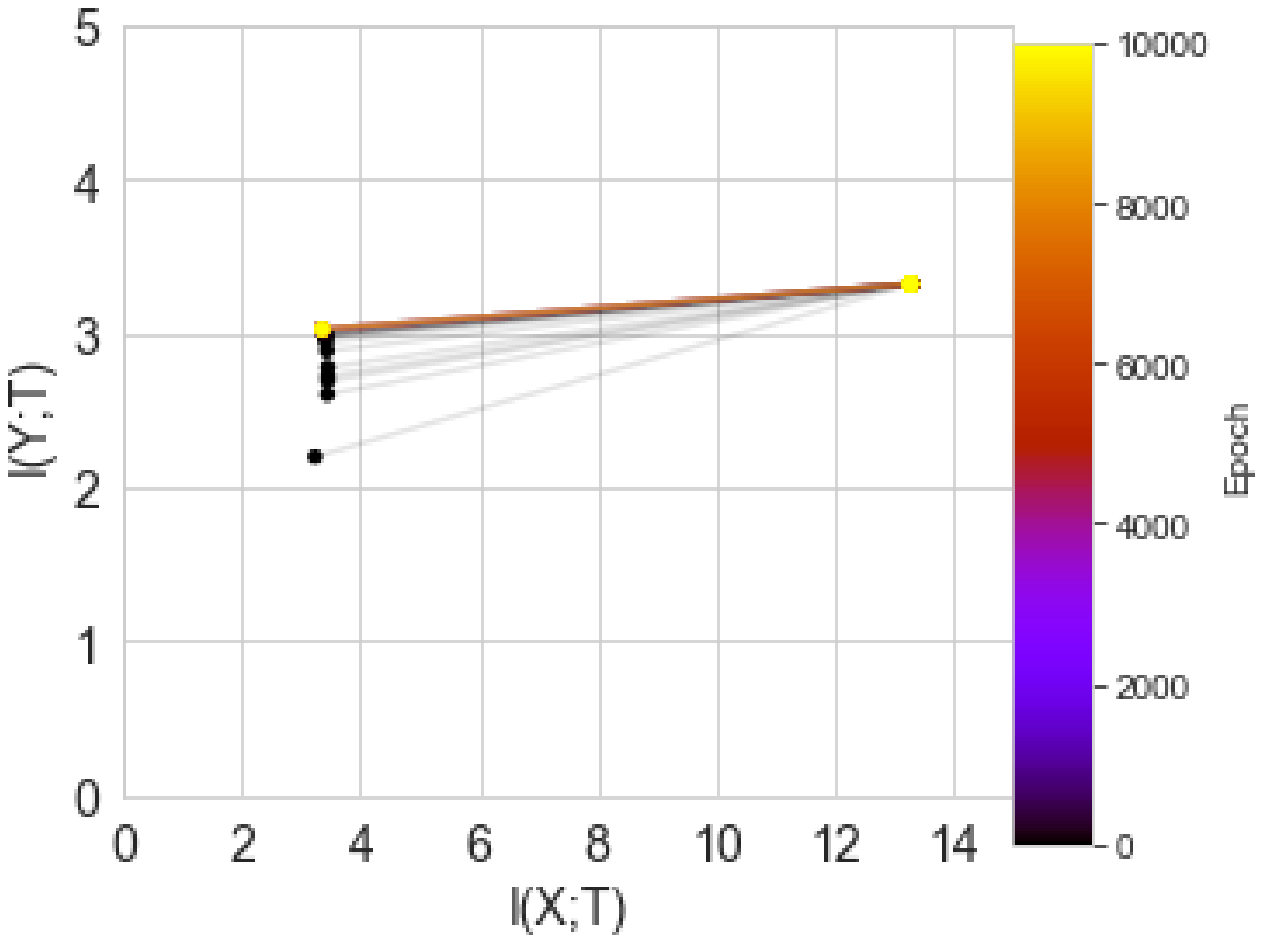}}
		\subfigure[IP on test data of Fashion-MNIST ]{
			\includegraphics[width=0.35\textwidth]{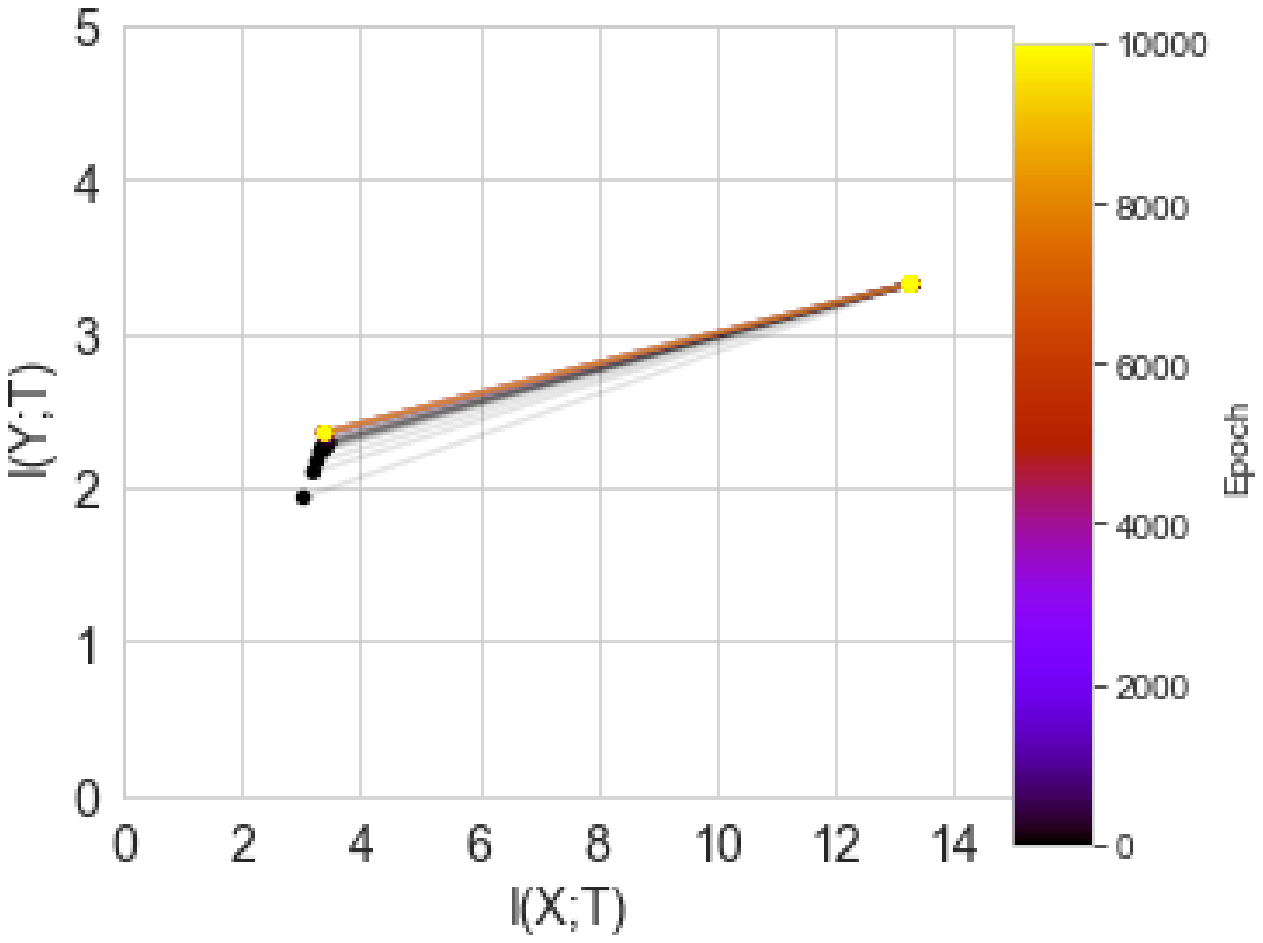}}
		
		\caption{Information plane for multi-fully connected layers on CNNs. }
		\label{fullyIP}
	\end{figure*}
	
	In this case, we use the fact that $H(T|X) =0$ (We treat every input data and the output of each layer as a vector variable respectively. Because of the high dimension of the input variable, each input data will correspond to a output of every layer. So the conditional entropy is equal to zero ). Then $I(T;X)$ and $I(T;Y)$ can be rewritten respectively as:
	
	\begin{equation}
	\begin{aligned}
	I(T;X) &= H(T)-H(T|X)
	\\&=H(T)
	\\&=-\sum_{i=1}^{n}p_i\log p_i
	\label{eq:I_X_T}
	\end{aligned}
	\end{equation}
	
	\begin{equation}
	\begin{aligned}
	I(T;Y) &= H(T)-H(T|Y)
	\\&=-\sum_{i=1}^{n}p_i\log p_i + \sum_{i=1}^{n}p_i H(T|Y=y_i)
	\label{eq:I_T_Y}
	\end{aligned}
	\end{equation}
	where $p_i$ is the probability that an activation output lands in the \textit{i}th interval.

	\subsection{Information Plane And Data Processing Inequality}
	Information plane (IP) shows the dynamic behavior of $I(Y;T)$ with respect to $I(X;T)$ ~\citep{shwartz2017opening}. They propose that the optimization of feedforward neural networks involve two phases, namely fitting phase and compression phase. In the fitting phase, the feedforward neural networks try to fit training samples into corresponding labels by increasing both $I(X;T)$ and $I(Y;T)$. In compression phase, the feedforward neural networks discard redundant information by reducing $I(X;T)$. Based on these statements, people will observe these two apparent phases on IP.

	Moreover, the MI among all feedforward neural networks layers form a Markov chain, which leads to Data Processing Inequality (DPI). It can be depicted as:
	\begin{equation}
	\begin{aligned}
	H(X)\geq I(X;T_0)\cdots I(X;T_{n-2})\geq I(X;T_{n-1})
	\\H(Y)\geq I(Y;T_0)\cdots I(Y;T_{n-2})\geq I(Y;T_{n-1})
	\end{aligned}
	\label{eq:DPI}
	\end{equation}
	where $T_0$ is input layer, $T_1 \cdots T_{n-2}$ are hidden layers, and $T_{n-1}$ denotes the final output layer.

	\section{Experiments and discussions}
	In order to investigate the impact of IB theory on CNNs , we perform a series of experiments on MNIST and Fashion-MNIST datasets which are two popular benchmarks on image classificaion.
	The training set consists of 60 000 ($28\times28$) gray-scale images, with 10 000 testing examples.
	For simplicity, we select 10,000 training samples randomly as training dataset and 10,000 test samples as test dataset. The networks are trained by using Adam algorithm and cross-entropy loss function with batch of 1000 samples. In addition, we set the learning rate as $10^{-3}$, and use $tanh$ activation except for final output layer with softmax. Our model is shown in Fig.~\ref{network}. The MI is evaluated on both training dataset and test dataset respectively. Therefore, in this case, $H(X)$ for both training dataset and test dataset equals to $\log_2{10^4}$. Then, we analyse the impact of some  crucial features such as convolutional layer width, network depth, kernel size and pooling layer on CNNs from view point of IB theory. Specifically, we discuss the compression phase on CNNs architecture. Our code is available on the Github.com \footnote{\href{https://github.com/mrjunjieli/IB_ON_CNN}{https://github.com/mrjunjieli/IB\_ON\_CNN}}.  The detailed information on these two benchmarks could be found at the official website: (http://yann.lecun.com/exdb/mnist/)
and (https://github.com/zalandoresearch/fashion-mnist).

	\subsection{Convolutional Layer Width}
	The convolutional layer width is crucial on the way to understand representation power of neural networks. To study the effect of convolutional layer width from the perspective of IB theory, we train 4 different CNNs with various of convolutional layer widths (number of channels).
	
	Fig.~\ref{widthMI} shows the $I(Y;T)$ and $I(X;T)$ paths on these networks during the training and test phase. By using DPI we introduce earlier, the theoretical upper bound of $I(X;T)$ of each layer is $H(X)$. Similarly, the theoretical upper bound of $I(Y;T)$ equals to $H(Y)$. Therefore, in this figure,  we observe the MI on all convolutional layers reach the upper bound which means they capture almost all information on input $X$ and label $Y$. This is due to we treat the whole image as a single variable, then all images are basically different. So according to Eq.~\ref{eq:I_X_T}, $I(X;T)$ can be represented by $H(T)$. Moreover, $H(T)$ is equal to  $\log_2{10^4}$ i.e. $H(X)$. In the same view, $I(Y;T)$ on convolutional layer is closely equal to $H(Y)$.
	
	For final output layer, the starting value of $I(X;T)$ and $I(Y;T)$ increase apparently with the expending of width. And also, with wider convolutional layer, the model reaches the upper bound faster. So wide CNNs can perform better with less training epochs. Specifically, in panel (c) and (d), we observe larger maximum value of $I(Y;T)$ for final output layer with increasing of width. Based on these observations, we believe that wide network is capable of capturing more information, which is beneficial to have better generalization.

	\subsection{Kernel Size and Network Depth}
	
	Ref.~\citep{szegedy2016rethinking} points out network with a large kernel size can be replaced by a deep network with small kernel size. From the perspective of information theory, how does the kernel size and depth affect MI in CNNs? We evaluate MI with various choices of depth and kernel size. By comparing  information paths in Fig.~\ref{Kernel} and Fig.~\ref{Depth}, we find that both larger kernel size and deeper network can promote the starting value of $I(X;T)$ and $I(Y;T)$ on the final output layer, which implies network capture more information with less training.  However, if we continuously increase kernel size or depth , the starting point cannot increase anymore or even becomes worse. So we propose that the larger kernel size and depth can drive network capturing more information with less training epochs. But over-large kernel size and depth will need more training epochs to capture the same amount of information. Furthermore, unlike convolutional layer width, in Fig.~\ref{Kernel} (e), (f), (g), (h) as well as in Fig.~\ref{Depth} (c) and (d), we observe that they all reach the same maximum value of MI for final output layer, which implies that a small kernel size and shallow depth are good enough to have a better generalization performance in these simple cases.

	\subsection{Pooling Layers and Multi-Fully Connected Layers}
	
	CNNs almost always include some forms of pooling layer such as max pooling and average pooling etc. The pooling layer always discards parts of data in order to improve the generalization and reduce computational complexity. In order to investigate the role of the pooling layer, we take the max pooling as example. In Fig. \ref{pool}, we observe that the curves of MI grow in different ways. Panel (c) and (d) show the networks with pooling layer reach a little bit larger value than networks without pooling layer, which implies that pooling layer is beneficial to have better generalization.
	
	We also design different CNNs with multi-fully connected layers to study whether double-sided saturating nonlinearities like $tanh$ yield compression phase in CNNs. (Ref.~\citep{saxe2019information} propose that  double-sided saturating nonlinearities yield a compression phase while linear activation function can not). So, we use a network with 5 convolutional layers (convolutional layer width=3-3-3-3-3 and kernel size=3-3-3-3-3) and 4 fully connected layers (500-1024-500-10). Fig. \ref{fullyIP} shows these layers information plane (IP) paths. From this we observe that all convolutional layers and fully connected layers (except for final output fully connected layers) converge to a point. In addition, the MI of final output layers grow during  both training phase and test phase. So, we observe there are no compression phase occurs in the CNNs.

	\section{Conclusion and future works}
	Information bottleneck theory provides a interesting analytic tool to explore the inner behavior of deep neural network, and based on this, people try to understand why deep learning works well. Along this way, this paper tries to extend the study to CNNs and investigate how the fundamental features have impact on the performance of CNNs. Based on our cases, we summarize some key observations and draw conclusions as:
	
	1. Convolutional layers can capture almost all information on input towards label. The MI between convolutional layers and input/output keep close to their upper bound.
	
	2. Wide convolutional layers network is able to improve the generalization performance. Furthermore, wide network need less training epochs to reach its optimal performance than narrow one.
	
	3. In general case, larger kernel size and deeper architecture drive network capturing more information with less training epochs. But, an over-large kernel size and over deep architecture will need much more training epochs to capture the same amount of information. This implies that people should balance the kernel size and network depth while design the deep architecture. Furthermore, it shows us the extremely deep neural network is probably not the right way to do deep learning.

	4. In some simple cases, our results also reveal that there is no compression whether in convolutional layers or fully connected layers, even using double-sided saturating nonlinearities in CNNs. Hence, we tend to think the compression probably happen in some specific cases, but not a universal mechanism in deep learning, and further, the relationship between it and generalization needs more experimental verification.
	
	In the future work, we plan to verify the above conclusions on some more complicated datasets such as ImageNet, and on some more complicated deep architectures such as Generative Adversarial Networks (GANs). We believe it will provide more experimental evidence to verify the IB theory and help us to understand deep learning.

\begin{acknowledgements}
Our research is supported by the Tianjin Natural Science Foundation of China (20JCYBJC00500), the Science \& Technology Development Fund of Tianjin Education Commission for Higher Education (2018KJ217).
\end{acknowledgements}

%
%

\bibliographystyle{spmpsci}      
\bibliography{main}   

\section{Appendix}

In order to further verify our conclusions, we conduct additional experiments on the CIFAR-10 dataset ~\citep{krizhevsky2009learning}. This dataset consists of 60,000 32x32 colour images in 10 classes, with 6,000 images per class.  There are 50,000 training images and 10,000 test images. 

In this experiment, the whole \textbf{50,000} training images and 10,000 test images are selected as our training dataset and test dataset respectively, which is the only different setting from \textit{Experiments and discussion} section. Furthermore, because of the arithmetic of computing mutual information, we choose to average the image of three channels and turn it into a signal channel as input data. 

\begin{figure*}[htbp]
		\centering
		
		\includegraphics[width=0.8\textwidth]{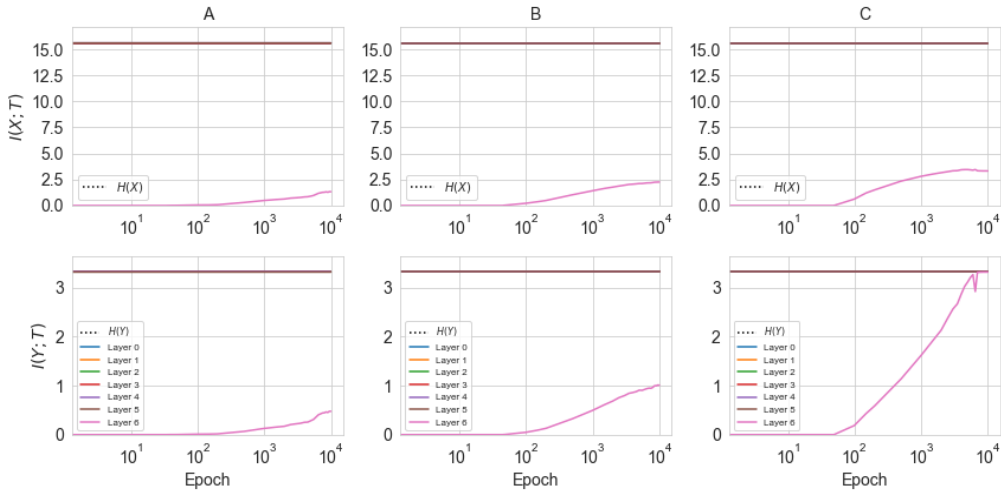}

		\caption{(Colored online) MI path on CNNs with different convolutional layer widths on training data of CIFAR-10. The convolutional layer width of 3 networks are (A) 3-3-3-3-3-3 (B) 6-6-6-6-6-6 (C) 12-12-12-12-12-12.}
		\label{cifarwidth}
\end{figure*}

\begin{figure*}[htbp]
		\centering
		
		\includegraphics[width=0.9\textwidth]{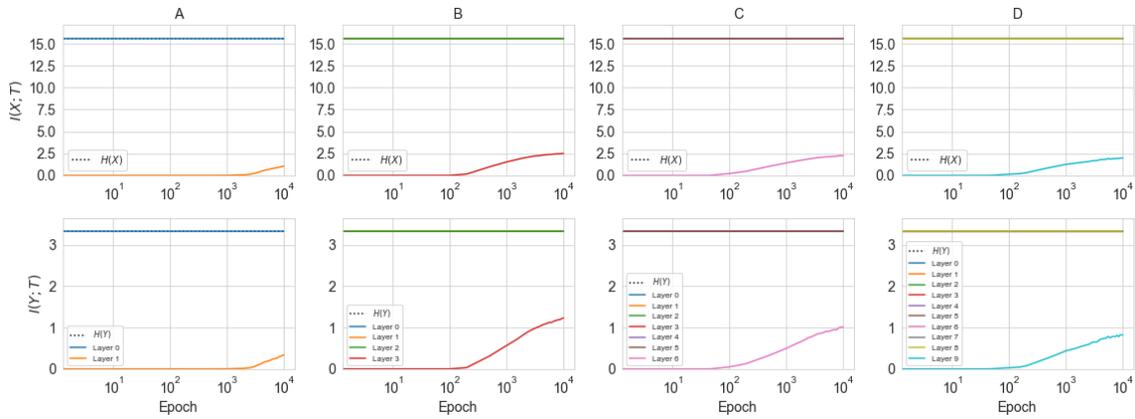}

		\caption{(Colored online) MI path on CNNs with different convolutional layer depths on training data of CIFAR-10. The  depth of 4 networks are (A) depth=2 (B) depth=4 (C) depth=7 (D) depth=10. All these networks width are set to 6.}
		\label{cifardepth}
\end{figure*}

\begin{figure*}[htbp]
		\centering
		
		\includegraphics[width=0.5\textwidth]{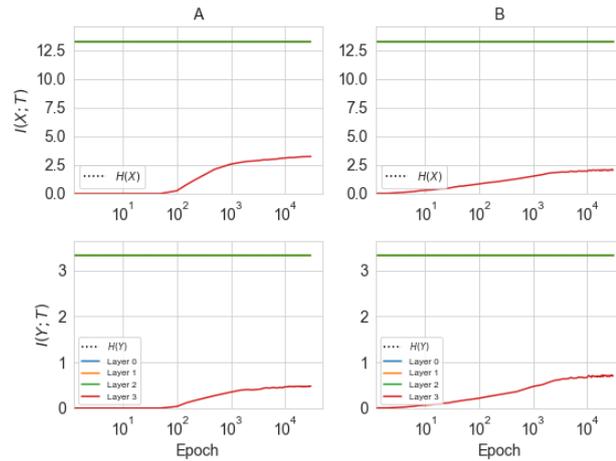}

		\caption{(Colored online) MI path on CNNs with pooling layer on test data of CIFAR-10. (A)without pooling layer (B)with pooling layer. The width of convolutional layers are both 6-6-6 and the kernel size is fixed to 3x3. Moreover, the MaxPooling2D layer is added after the layer 1 and the pooling size is set to 2x2.}
		\label{cifarpooling}
\end{figure*}

The Fig.~\ref{cifarwidth} and Fig.~\ref{cifardepth} show the MI with different widths and depths on training data respectively. Fig.~\ref{cifarpooling} shows the MI with pooling layer on test data. These results offer more proof about the IB theory.

\end{document}